\documentclass{article}
\usepackage{ijcai24}

\usepackage{times}
\usepackage{soul}
\usepackage{url}
\usepackage[utf8]{inputenc}
\usepackage[small]{caption}
\usepackage{graphicx}
\usepackage{amsmath}
\usepackage{amsthm}
\usepackage{booktabs}
\usepackage{algorithm}
\usepackage{algorithmic}
\usepackage[switch]{lineno}
\usepackage{amsfonts}
\usepackage{multirow}
\usepackage{makecell}
\usepackage{pifont}
\usepackage[dvipsnames]{xcolor}

\definecolor{cvprblue}{rgb}{0.21,0.49,0.74}

\usepackage[pagebackref,hidelinks,breaklinks,colorlinks,citecolor=cvprblue]{hyperref}

\title{Let Storytelling Tell Vivid Stories: An Expressive and Fluent \\ Multimodal Storyteller}
\author{
Chuanqi Zang$^1$\
, Jiji Tang$^2$\
, Rongsheng Zhang$^2$\
, Zeng Zhao$^2$\
, Tangjie Lv$^2$,\\
Mingtao Pei$^1$\
, Wei Liang$^1$\\
{\small$^1$School of Computer Science and Technology, Beijing Institute of Technology}\\
{\small$^2$Fuxi AI Lab, NetEase Inc.}\\
{\small jnchuanqizang@gmail.com},
{\small \{tangjiji01,zhangrongsheng,hzzhaozeng,hzlvtangjie\}@corp.netease.com},
{\small \{peimt,liangwei\}@bit.edu.cn}
}
\begin{document}

\twocolumn[{%
\renewcommand\twocolumn[1][]{#1}%
\maketitle
\begin{center}
    \centering
    \captionsetup{type=figure}
    \includegraphics[width=\textwidth]{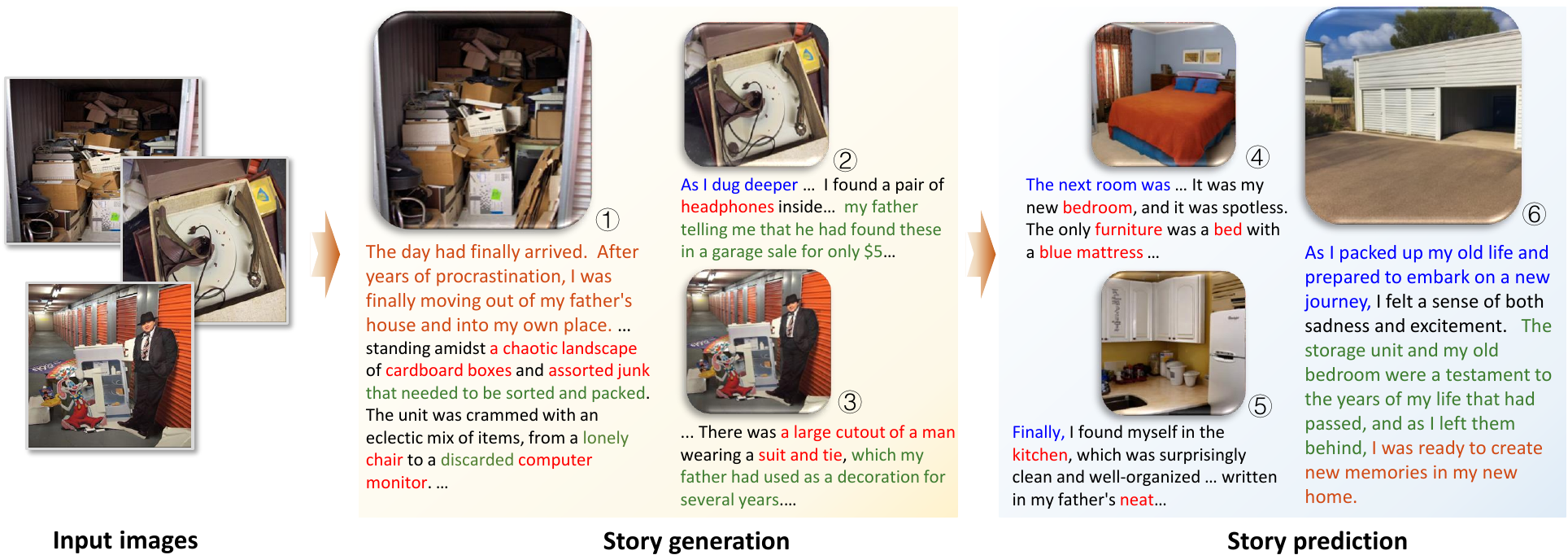}
    \captionof{figure}{In this work, we unify story generation and story prediction as storytelling and propose a novel pipeline for this task. Observing 3 images of a story, our work initially generates a vivid story across a reasonable storyline based on the factual events occurring in the images. Then, aligning this storyline, our work imagines and forecasts subsequent story developments, presenting them through textual descriptions and accompanying visual representations. The results exhibit comprehensive story semantics, both in story expressiveness (\textcolor{Bittersweet}{integrality}, \textcolor{OliveGreen}{interestingness}, \textcolor{red}{correlation}) and multiple story plots \textcolor{blue}{consistency}.}
    % \textcolor{Bittersweet}{integrality}, \textcolor{OliveGreen}{interestingness}, and \textcolor{blue}{consistency}, with promising multi-modal \textcolor{red}{correlation}.
    
    % collectively refer to story generation and story prediction as storytelling. With just a few images of a story (left), our work can first generate a reasonable and vivid story based on the explicit fact of images and then predict consistent story plots along the previous storyline with imagination, displayed in text and images (right). 
    % The results exhibit rich story semantics, reasonable story connection, and extended storylines, as well as illustrations with high fidelity to the plot text and the observed visual style.
    
\end{center}%
}]

\begin{abstract}
Storytelling aims to generate reasonable and vivid narratives based on an ordered image stream. The fidelity to the image story theme and the divergence of story plots attract readers to keep reading. Previous works iteratively improved the alignment of multiple modalities but ultimately resulted in the generation of simplistic storylines for image streams. In this work, we propose a new pipeline, termed \textbf{LLaMS}, to generate multimodal human-level stories that are embodied in \textbf{expressiveness} and \textbf{consistency}. Specifically, by fully exploiting the commonsense knowledge within the LLM, we first employ a sequence data auto-enhancement strategy to enhance factual content expression and leverage a textual reasoning architecture for expressive story generation and prediction. Secondly, we propose SQ-Adatpter module for story illustration generation which can maintain sequence consistency. Numerical results are conducted through human evaluation to verify the superiority of proposed LLaMS. Evaluations show that LLaMS achieves state-of-the-art storytelling performance and 86$\%$ correlation and 100$\%$ consistency win rate as compared with previous SOTA methods. Furthermore, ablation experiments are conducted to verify the effectiveness of proposed sequence data enhancement and SQ-Adapter.

\end{abstract}

% 故事性，文本故事性
% 图片故事性。图片可视化一致性
% 现有storytelling的工作都在关注于如何对齐一个高连贯性但是低故事性的文本
% 我们认为，基于LLM在文本生成上得能力，这个任务可以做得更好。需要定义一个新的研究范式来推进这个任务。
% 数据上，受llava在图文对齐任务上得启发，我们可以用auto的形式，做更高质量的数据。与之前任务不同的是，我们需要额外关注数据的序列一致性
% 任务上，除了要保证数据的故事性，还需要保证多模态的一致性，尤其是现有的图像生成工作都是基于单图的提示，在序列任务中，未来图像的生成需要基于之前图像的语境与风格。

% 总体上，我们的方法做到了高故事性，与阅读上的连贯性与承接性。生成的数据更贴合我们对这个任务的预期与目标。

% factual descriptions, imaginary concepts 
% coherence, reasonability, narrative, multimodal correlation, integrality
% informative commonsense, visual relationship among objects
% diversity, informativeness
% data-efficient

\section{Introduction}
Do you want to read a vivid story book begin with your predetermined beginning? How about changing the ending of the comic to another possible plot (bad ending to happy ending)? Rendering such application scenes is a general storytelling task that generates cohesive narratives represented as language and image, conditioned on a few images. The storytelling task is challenging for it requires tying disparate moments and synthesizing a coherent multimodal story across time so that they seamlessly blend into a complete story.
% Early works~\cite{park2015expressing, huangferraro2016visual, Xu2021ImagineRA} focus on the intersection of multimodal data and reason the story from a single domain. MiniGPT-5~\cite{Liu2023VisualSW} generalizes the framework into multimodal generation, improving the scope of storytelling applications, such as comic creation, picture book creation, and picture story creation
% which requires not only the observed multi-image understanding in generating a story but also reasonable temporal reasoning aligning the past story. 

Previous storytelling methods \cite{park2015expressing,huangferraro2016visual,Xu2021ImagineRA} employ the vision-to-text model to generate stories based on semantic concepts of visual features and event correlations among images. Despite elaborate designs of these models for alignment in either the whole story or specific objects, the result of the storytelling task is frustrating due to poor narrative and diversity. 

Such poor storytelling results can be attributed to two aspects. First, they record events in multiple images by brief descriptions based on restricted visual elements, disregarding rich interactive information between elements and the expended imagination. Current multi-modal LLM, such as LLaVa~\cite{liu2023llava}, consists of an image encoder with strong image understanding ability and an LLM decoder with powerful text description ability that can describe visual objects with few oversights. However, due to the lack of multi-image understanding ability, LLaVa struggles with generating sequential consistent stories. Secondly, current single-modal story generation pipelines limit the diverse expression of storytelling.  MiniGPT-5~\cite{zheng2023minigpt5} can generate results in both text and pictures but produces poor expressiveness and consistency. In order to immerse ourselves in a story, we usually prefer interesting plots, supplemented by consistent visual imagery.

Firstly, the storytelling ability of existing models is restricted by current low-quality training data. However, urgently required high-quality data previously relied heavily on extensive manual annotation, which is difficult for long story annotation. To this end, we propose a new pipeline, called \textbf{L}arge \textbf{L}anguage model \textbf{a}ssisted \textbf{M}ultimodal \textbf{S}toryteller (\textbf{LLaMS}).  In the pipeline, we propose a unique \textbf{sequential data quality enhancement strategy} to rewrite the story automatically. The enhanced data integrates richer visual semantics and story evolution logic that motivates the following textual storytelling model to generate high-expressive stories. 
% take a meaningful step further toward the goals of the storytelling task. Benefiting from the emergence of pre-trained base models, we
% , relying on the commonsense expression and logical reasoning capabilities of LLM. 
% Considering the mechanical "adventure plots" template commonly used in LLM, we expand the story base with supervised storylines.

Secondly, humans prefer to read multimodal stories with visual consistency in sequence. For image style controlling, recent advanced methods, such as IP-Adapter~\cite{ye2023ip}, have demonstrated that a single image can effectively control the generated content of the final image. However, they struggle to capture sequence common styles (such as season, scene, etc.) in the human-provided images. In our LLaMS, we propose a \textbf{SQ-Adapter} module to map variable-length story visual information into a learnable latent code. It integrates the style information of multiple images to provide prompt information during the image generation process, which eventually ensures visual consistency. 
% To generate high-expressive stories, we propose a two-stage architecture including an image-to-text model and a text-to-image model to express stories in sentences and illustrations. During the text-to-image stage, consistent image style is essential for human-preferred stories. 
% To enhance visual consistency, we propose a parameter-saving adapter module to generate images consistent with historical image styles. 
% After deep self-attention correlation modeling, the latent code can learn historical style preferences. During story prediction, our adapter module follows textual instructions as well as latent code, generating multimodally coherent and visually consistent results.
% We leverage a deep integration model to map variable-length story visual information into a learnable latent code.  to control the style of the future story image. With the predicted text plots, our LLaMS can generate highly expressive and consistent results.

We summarize our contributions as:

(1) We propose a novel pipeline (LLaMS) for the storytelling task to generate human-preferred stories by enhancing expressiveness and maintaining consistency. LLaMS achieves the SOTA performance in various metrics.

(2) We introduce a multi-modal storytelling architecture with an automatic sequence data enhancement strategy, effectively improving the expressiveness of stories. 

(3) We are also the first to propose a style consistency adapter model that captures the common features of multiple images and improves the overall story consistency. 
% We propose to take expressiveness and consistency as the core of the storytelling task, expressing a readable story result. 
% (3) We deliver extremely impressive results and perform comprehensive evaluations in two datasets using automated evaluators and humans.

\section{Related work}
\subsection{Storytelling Task}
The storytelling task was first proposed by Park and Kim~\cite{park2015expressing}, aiming to generate consistent narratives rather than multi-image captions. Huang et al.~\cite{huangferraro2016visual} released a visual storytelling (VIST) dataset that incrementally attracts efforts on stories-in-sequence. According to this benchmark, visual coherence strengthening method~\cite{Liu2017LetYP,Kim2018GLACNG}, commonsense-driven methods~\cite{Yang2019KnowledgeableSA,Hsu2019KnowledgeEnrichedVS}, and external knowledge-based method~\cite{Xu2021ImagineRA,chen2021commonsense} were proposed to generate a coherent story with ordered images, using RNNs model~\cite{rumelhart1986learning,hochreiter1997long}, reinforcement learning-based model~\cite{chen2017multimodal}, or transformer model~\cite{vaswani2017attention}. 

Except for text story generation, our work integrate story prediction into storytelling task by a multi-modal generation pipeline, broadening the application scenarios. Fully leveraging the capabilities of pre-trained models, our pipeline generate superior stories compared to existing works.
% Supported by pre-trained models, keypoint annotations-assistant method~\cite{Liu2023VisualSW}, commonsense-driven methods~\cite{Yang2019KnowledgeableSA} achieve promising results in story generation.

\subsection{Multimodal Large Language Models}
Integrating visual understanding into the LLM attracts growing research groups to expand the application scenarios of LLM. Frozen~\cite{tsimpoukelli2021multimodal} was an early work that achieved excellent zero-shot multimodal generation ability using a gated cross-attention aligning multimodal feature followed by LLM. BLIP-2~\cite{li2023blip}, MiniGPT-4~\cite{zhu2023minigpt} and LLaVA~\cite{liu2023llava} utilized the ViT~\cite{dosovitskiy2020image} and LLaMa~\cite{touvron2023llama}/Vicuna~\cite{vicuna2023}/ as powerful base models and leveraged high-quality image-text pairs, achieving amazing single image understanding performance. NextGPT~\cite{wu2023nextgpt} is trained on any-to-any multimodal pairs and showed improved instruction-following learning capabilities.

In order to generate multi-modal stories, MiniGPT-5~\cite{zheng2023minigpt5} first proposed to use voken~\cite{tan2020vokenization} instead of token to represent the output features of LLM, which can be used for text and image generations simultaneously. But they still struggle in story expression and consistency. In our work, we leverage pre-trained LLM to improve the quality of the dataset and utilize an architecture to accomplish story generation and prediction. Our pipeline provides novel solutions to storytelling tasks, improving the expressiveness and the consistency of multimodal results.
% These approaches take advantage of the capabilities of LLM to understand visual semantic content and reason explicitly derived information. 
% They train billions of parameters using billions of image-text pairs which also demonstrate the benefits of massive data in multimodal generation.

% MiniGPT-5~\cite{zheng2023minigpt5} firstly proposes using voken instead of token to represent the output feature of LLM, which is ultimately used for multimodal story generation based on previous reviews. 
% In our work, we propose a new framework for sequence-correlated multimodal story generation. Our framework can generate long-form stories with richer storytelling and consistency. 
% In addition to predicting future stories based on previous reviews, we can also generate diverse story scripts based only on images.

\begin{figure*}[htb]
\begin{center}
\includegraphics[width=17cm]{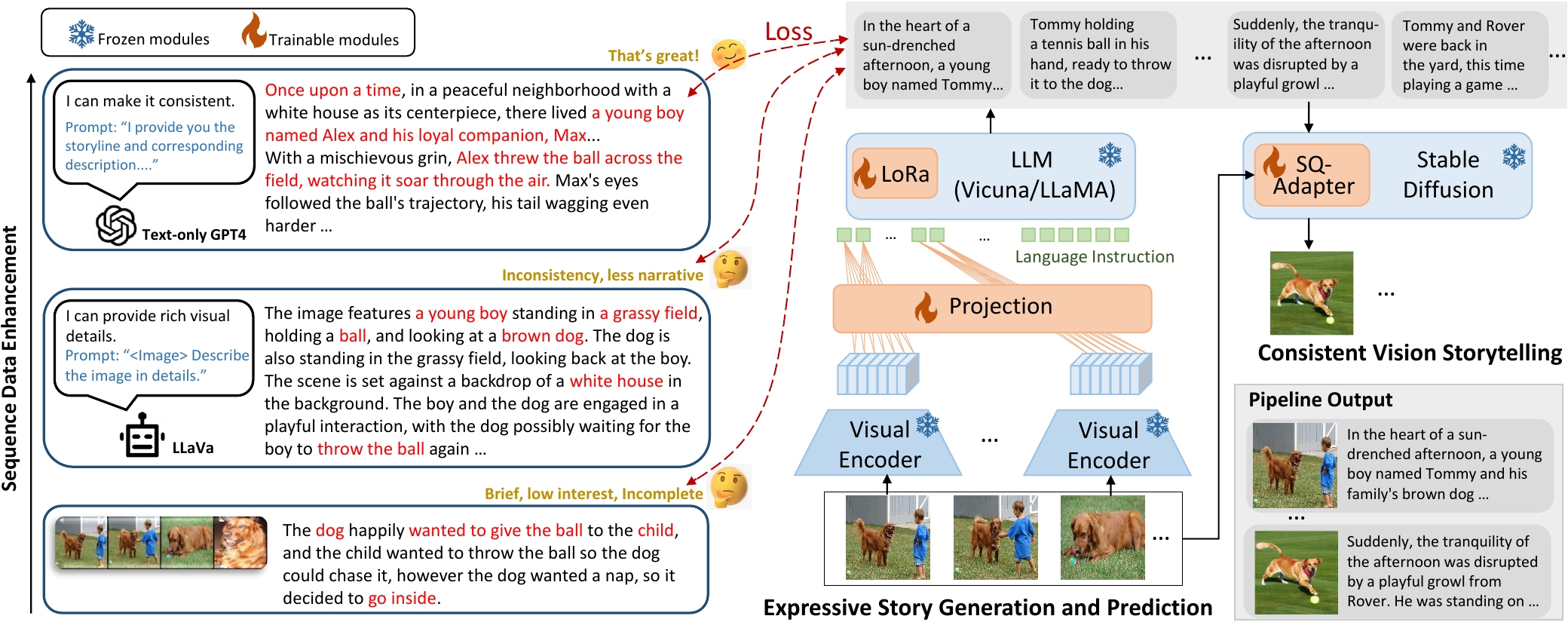}
\end{center}
\caption{Overview of Large Language model assisted Multimodal Storyteller (LLaMS) pipeline. Left is the process of sequence data enhancement from brief storylines to detailed descriptions to high-expressive (integral, interesting, correlated) and consistent stories. Right is the architecture for the storytelling task. During the inference stage, given 1$\sim$5 image of a story, we design an image-to-text model to generate textual stories based on images and a text-to-image model to generate style-consistent plots illustrations.}
\label{fig:data}
\end{figure*}

\section{Preliminary}
\subsection{Storytelling Formalism}
Given the inspired image stream $I=\{i_0, i_1,..,i_k\}$ with length $k$, the story generation task requires capturing the event flow in the image stream, generating a coherent fact-conditioned story $Y=\{y_0, y_1,...,y_k\}$. Based on the factual, story prediction task aims to infer a reasonable future multimodal story $F=\{\{i_{k+1}, y_{k+1}\}, ..., \{i_{N}, y_{N}\}\}$, where N is the number of story plot. These two storytelling tasks can be accomplished by two sub-tasks: image-to-text generation and text-to-image generation.

\subsection{Image-to-Text Caption Model}
LLaVa~\cite{liu2023llava} is the state-of-the-art image-to-text caption model for single-image understanding. Given an image $i$, LLaVa uses a pre-trained CLIP encoder~\cite{radford2021learning} to map $i$ to the image feature space. Then, they use a two-layer nonlinear Projection to convert the image feature to the LLM-preferred text embedding tokens $v \in \mathbb{R}^{L_v \times C}$. $L$ is the length of patch tokens and $C$ is the dimension of the feature embedding.
 Finally, LLM $G$ integrates these converted tokens and language instruction tokens to generate caption. The LLaVa captures image semantics and responds to human instructions, which can be formalized by:
\begin{equation}
\begin{gathered}
\hat{y}=G\left(\left[v ; E_{\text {instruct }}\right]\right), v=MLP\left(CLIP_I(\mathrm{i})\right),
\end{gathered}
\end{equation}
where $\hat{y}$ is the generated caption, $E_{instruct} \in \mathbb{R}^{L_i \times C}$ means the word embedding of human instructions, $\left[;\right]$ denotes concatenation operation, $MLP$ is the two-layer Projection. 
% $W$ and $G$ are trainable models in LLaVa architecture.

% LLaVa expands the description by finetuning the pre-trained model with high-quality data.

% In this pipeline, $\boldsymbol{E}$ maps observed images to visual features $v$, and $\boldsymbol{M}$ maps the feature to text semantic tokens that LLM prefers. These tokens are sent to LLM $\boldsymbol{G}$ together with task instructions for answer generation $y$. The probability of answer generation can be represented by:
% \begin{equation}
% p=\prod_{i=1}^n p_\theta\left(\boldsymbol{y}_i \mid \boldsymbol{i}, \boldsymbol{y}_{<i}, O\right),
% \end{equation}
% where $\theta_m$ corresponds to the trainable parameters during training, $O$ is the prompt for the task.
% $\boldsymbol{\epsilon} \sim \mathcal{N}(0, \mathbf{I})$
\subsection{Text-to-Image Generation Model}
Stable Diffusion (SD)~\cite{rombach2022high} is a commonly used generative model that generates an image from a variable sampled from a Gaussian distribution. When using text $y$ as a semantic controller, SD extracts the text embedding $c$ by the pre-trained CLIP model, recorded as $c = CLIP_T(y)$. U-Net $\epsilon_\theta$ gradually denoises noisy features $x_t$ referring text embedding $c$ with multiple steps. The gradual diffusion process can be represented by
\begin{equation}
x_{t-1}=\frac{1}{\sqrt{\alpha_t}}\left(x_t-\frac{1-\alpha_t}{\sqrt{1-\bar{\alpha}_t}} \epsilon_\theta\left(x_t, c, t\right)\right)+\sigma_t z, 
\end{equation}
where $z$ is sampled from Gaussian noise, $t\in\left[0, T\right]$ represents the diffusion process time step, $\alpha_t, \sigma_t$ are used to control the noise process and sample quality, respectively. The denoised results eventually decode to the real image.
% SD is composed of a text encoder $\boldsymbol{B}$, a denoising U-Net $\epsilon_\theta$, and an image decoder $\boldsymbol{D}$. 
% The training objective of controllable diffusion models is to predict a denoised variant of input, defined as a variant of the variational bound:
% \begin{equation}
% \mathbf{E}_{\boldsymbol{i}_0, \boldsymbol{\epsilon} \sim \mathcal{N}(\mathbf{0}, \mathbf{I}), \boldsymbol{c}, t}\left\|\epsilon-\epsilon_\theta\left(\left(\alpha_t \boldsymbol{i}_0+\sigma_t \boldsymbol{\epsilon}\right), \boldsymbol{c}, t\right)\right\|^2, c=CLIP_T(y),
% \end{equation}
% \begin{equation}
% x_{x-1} = \alpha_t\left(x_t-\sigma\epsilon_\theta\left(x_t,c,t\right)\right), c=CLIP_T(y),
% \end{equation}

% \section{Method}
% Multimodal storytelling task requires higher-quality data and an effective framework for training. We will introduce our work on high-quality data generation in Sec.~\ref{data} and our simple but effective framework in Sec.~\ref{LLAST}
\section{Method}
In this paper, we formulate a novel and efficient sequential vision-language pipeline to generate highly expressive and consistent narratives. As illustrated in Fig.~\ref{fig:data}, we achieve that in two stages: expressive story generation and prediction stage and consistent vision storytelling stage.
% : (1) We automatically enhance the story dataset with pre-trained upstream models. (2) We design a two-stage architecture that can predict textual stories based on factual visual concepts in observed images and then generate illustrations consistent with the style of conditioned images.

\subsection{Expressive Story Generation and Prediction}

\subsubsection{Sequence Data Enhancement}\label{data}
% The community has been working on generating stories based on VIST dataset~\cite{huangferraro2016visual}. However, previous methods generate poor plot content for the image sequence, partially because low-quality image-text alignment content and high-degree-of-freedom storylines increase the difficulty of the multi-modal storytelling task. Careful annotation in story creation and polishing is highly expensive and hard to scale to millions of storytelling data.

Recent storytelling works use the VIST dataset as their core dataset. We illustrate a specific case in Fig.~\ref{fig:data} (bottom left). The story begins with ``wanted to give the ball" and ends with ``go inside". The storyline is simple and lacks expression in protagonists, story scenes, and activity content. The connection between plot segments is obscure and requires readers to take the initiative to figure it out.

% In order to break the constraints of low-quality data on the storytelling task, 
% % Previous methods produce poor plotting content for image sequences, an extremely important factor being low-quality image-text alignment data. Inspired by the success of data enhancement in the image caption task with GPT models, 
% we leverage the pre-trained multi-modal caption model and GPT-4 for sequence data enhancement in two stages. One sample is shown in Fig.~\ref{fig:data} (bottom left).

% To collect rich semantic visual information in an image sequence, details of each image need to be recorded in the first stage. 
In this work, we first collect more story details about the storyline by LLaVa~\cite{liu2023llava}. We prompt LLaVa to focus on details of visual semantics and describe highly deterministic parts of them in language, including event scenes, object characteristics, and interactions. As shown in Fig.~\ref{fig:data} (left middle), rich fine-grained semantics are explicitly expressed, such as ``a grassy field", ``white house", ``young boy", ``brown dog", and ``throw the ball", improving the correlation between text and images. However, such descriptions lack the consistency between different plots and interestingness for continued reading.
 % to create image-faithful caption data.
% Text description per image increased from $\sim$10 words to over 70 words.
% It can adequately describe explicit objects and implicit interactions in images, using over 70 words per image.

Motivated by this, we use post-processing to integrate these independent contents into a complete story. A straightforward way is to use language-only GPT-4 to tie contents through its common sense knowledge. However, this way is restricted by the choice of storyline. LLM prefers ``adventure story" and lacks story diversity in deductions and twists. In our pipeline, we use short storyline references provided by the original dataset as specific guidance apart from detailed descriptions. We prompt the GPT-4 to generate an integrated story, keeping the original storyline and imagining the plot content based on corresponding descriptions. As shown in Fig.~\ref{fig:data} (left top), the new enhanced data becomes an integrated and vivid narrative, expressed in ``Once upon a time", ``Alex threw the ball", and ``soar through the air".

Besides, lack of regulatory restrictions, the output formats of GPT are diverse. The enhanced story content may not correspond accurately to the images in content and order, making the multi-modal model struggle to converge. In this case, we restrict the paragraph length and order of output in the prompt to be consistent with the image order and directly drop illegal data. We eventually obtain 16k enhanced training data from 40k samples in the VIST training set. Each plot expands from the original $\sim$10 words to over 70 words.

% We select two datasets for sequence data enhancement: VIST~\cite{huangferraro2016visual} and MovieNet~\cite{huang2020movie} for two perspectives. \textbf{1) Narrative:} The source data contains diverse storylines that can be directly learned by LLaMS and guide the LLaMS to learn human preferences. \textbf{2) Multimodal consistency:} Multimodality is important for data representation, both for reading and training. Corresponding multi-modal data can speed up model convergence.

\subsubsection{Textual Storytelling Model}
% In this stage, our architecture adopts the LLaVa~\cite{liu2023llava} training paradigm to align vision and language data. 
% However, the LLaVa only focuses on the single image caption task, hardly handling multiple images and sequence data. 
% We consider an effective training strategy to train a sequence-sensitive.
% As illustrated in Fig.~\ref{fig:data} (right), similar to LLaVa architecture, we leverage the pre-trained Visual Encoder to process the visual semantics and LLM to understand basic elements of vision and reason about extended storylines. A fully connected Projection network is used to map visual features to the language representation. 
With highly expressive story data, we use LLaVa as a basic model to generate corresponding story plots based on provided images and predict future story plots. For the sequential correlation integration of multiple images in storytelling, we use two training strategies to complete it.

\textbf{Story Generation.} Given image sequence $I$, we use image encoder $CLIP_I$ to encode each image to visual features, recorded as $\tau_{0:N}=CLIP_I \left(i_{0:N} \right)$. $\tau_{0:N}$ are then projected to textual tokens by a two-layer Projection. The results are connected in sequence dimension and sent to LLM in image stream order, recorded as $ v_{0:N} \in \mathbb{R}^{N\cdot L_v \times C}$. We ultimately rely on LLM $G$ to infer story plots that correspond to images in paragraph order, which can be represented by

% ordered with story plots order and sent to LLM preferred embedding tokens $X= \{x_0, x1,..., x_k\}$ by the projection module. Visual tokens $X$ are sent to LLM in sequence order for text story generation. Projection $\boldsymbol{M}$ and LLM $\boldsymbol{G}$ are trained by predicting the entire text story, the training objective is defined as

\begin{equation}
\begin{gathered}
\hat{y}^g_{0:N}=G\left(\left[v_{0:N} ; E_{\text {instruct }}\right]\right),
\end{gathered}
\end{equation}
where $N$ represents the story plot length, $\hat{y}^g_{0:N}$ represents generated multiple story plots.

% Such a naive training strategy is insensitive to the image sequence and can not predict the future story with reasonable imagination. We use additional two effective training strategies to train our model. 

% \textbf{Sequence Order Learning.} Considering that text paragraphs are aligned to images during sequence data enhancement, we can split a plot form story with the corresponding image. During training, we will change the image order of samples with a $20\%$ probability, and the corresponding story text will follow the new order. The new train sample can be represented by new image-text pair, ($I^o$, $Y^o$). We train the modal with new pairs by
% \begin{equation}
% \mathcal{L}^o\left(\theta_m, \theta_l\right)=\prod_{i=1}^N p_\theta\left(\boldsymbol{y}^o_i \mid \boldsymbol{i^o}_{<=N}, \boldsymbol{y}^o_{<i-1}, O\right).
% \end{equation}
% After multiple iterations, the plot change strategy guides the model to be sensitive to sequence order. 

\textbf{Story Prediction.} Except for training the story generation task (N images to N plots), we also train the story prediction task (k images to N plots) in a unified model, which utilizes the sequence inference capabilities of LLM, denoted as
% \begin{equation}
% \mathcal{L}^p\left(\theta_m, \theta_l\right)=\prod_{i=1}^N p_\theta\left(\boldsymbol{y}_i \mid \boldsymbol{i}_{<=k}, \boldsymbol{y}_{<i-1}, O\right).
% \end{equation}
\begin{equation}
\begin{gathered}
\hat{y}^p_{0:N}=G\left(\left[v_{0:k} ; E_{\text {instruct }}\right]\right), k \in \left[1,N-1\right],
\end{gathered}
\end{equation}
where $k$ is sampled from discrete uniform distribution during training that enables the model to flexibly handle image streams with different lengths.
% where $k \in \{1,2,...,N-1\}$ denotes the number of provided images.
% For leaking images in predicting, we simply zero out the CLIP feature to pad the image token.

During the training stage, we jointly train our model with two strategies, represented by 
\begin{equation}
\mathcal{L}_{unite}=\lambda_g CE\left(\hat{y}^g_j, {y}_j\right)+\lambda_p CE\left(\hat{y}^p_j, {y}_j\right),
\end{equation}
where $\lambda_g$ and $\lambda_p$ are weight factors, and $CE$ means cross-entropy. In this way, our method can generate stories or predict stories based on variable-length image sequences. The model will degenerate to generation architecture if $\lambda_p=0$. The story generation optimization is necessary which ensures the convergence of the model on multi-modal alignment.
% In summary, story generation optimization ensures the convergence of the model on multi-modal alignment and story prediction optimization enables the model to predict story plots based on variable-length images.

\begin{figure}[ht]
\begin{center}
\includegraphics[width=7cm]{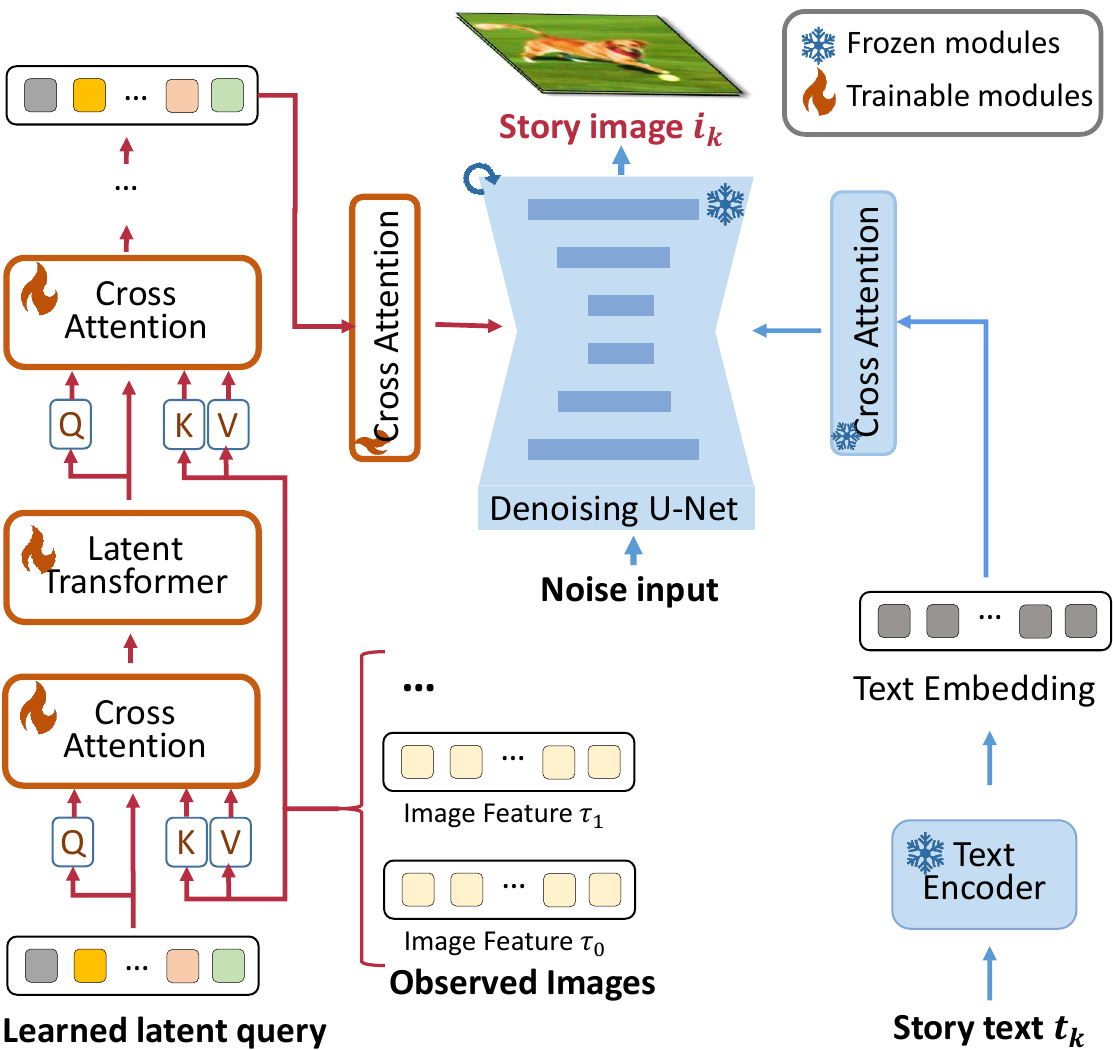}
\end{center}
\caption{We propose the SQ-Adapter for consistent vision generation. 
A learned latent query scales high-dimensional inputs without length limitation to a fixed dimension, controlling the generated image style. The SQ-Adapter is trainable with only a few parameters, shown in red. The frozen stable diffusion module is shown in blue.}
\label{fig:sq-adapter}
\end{figure}

\begin{table*}[ht]
  \centering
  \caption{Human evaluation results for story generation (5 images to 5 plots) and story prediction (3 images to 5 plots with 2 images). For clarity, we use ``Win+" and ``Lose+" to represent Win+Tie/2 and Lose+Tie/2, respectively. The raw results can be found in the supplementary.}
  \scalebox{0.95}{
    \begin{tabular}{ccccccccccc}
    \toprule
    \multirow{2}[4]{*}{Task} & \multicolumn{1}{c}{\multirow{2}[4]{*}{\thead{Models\\ (LLaMS vs. *)}}} & \multicolumn{1}{c}{\multirow{2}[4]{*}{\thead{LLM\\{}(Vicuna-13B)}}} & \multicolumn{2}{c}{Integrality} & \multicolumn{2}{c}{Interestingness} & \multicolumn{2}{c}{Consistency} & \multicolumn{2}{c}{Correlation} \\
\cmidrule{4-11}          &       &       & Win+  & Lose+ & Win+  & Lose+ & Win+  & Lose+ & Win+  & Lose+ \\
    \midrule
    \multicolumn{1}{c}{\multirow{6}[4]{*}{{\thead{Story Generation\\Task}}}} & AREL  & -     & \textcolor[rgb]{ 1,  0,  0}{0.89 } & 0.11  & \textcolor[rgb]{ 1,  0,  0}{0.91 } & 0.09  & \textcolor[rgb]{ 1,  0,  0}{1.00 } & 0.00  & \textcolor[rgb]{ 1,  0,  0}{0.85 } & 0.15  \\
          & RECO  & -     & \textcolor[rgb]{ 1,  0,  0}{0.92 } & 0.09  & \textcolor[rgb]{ 1,  0,  0}{1.00 } & 0.00  & \textcolor[rgb]{ 1,  0,  0}{1.00 } & 0.00  & \textcolor[rgb]{ 1,  0,  0}{0.93 } & 0.07  \\
          & LLaVa  & Vicuna-13B & \textcolor[rgb]{ 1,  0,  0}{0.98 } & 0.02  & \textcolor[rgb]{ 1,  0,  0}{1.00 } & 0.00  & \textcolor[rgb]{ 1,  0,  0}{1.00 } & 0.00  & \textcolor[rgb]{ 1,  0,  0}{0.67 } & \textcolor[rgb]{ 0,  .439,  .753}{0.33 } \\
          & NextGPT & Vicuna-7B  & \textcolor[rgb]{ 1,  0,  0}{0.62 } & \textcolor[rgb]{ 0,  .439,  .753}{0.38 } & \textcolor[rgb]{ 1,  0,  0}{0.89 } & \textcolor[rgb]{ 0,  .439,  .753}{0.11 } & \textcolor[rgb]{ 1,  0,  0}{0.88 } & \textcolor[rgb]{ 0,  .439,  .753}{0.13 } & \textcolor[rgb]{ 1,  0,  0}{0.99 } & 0.01  \\
\cmidrule{2-11}          & LLaMS-7B & Vicuna-7B  & 0.54  & 0.47  & 0.50  & 0.50  & 0.51  & 0.49  & 0.53  & 0.47  \\
          & G.T.  & -     & 0.34  & \textcolor[rgb]{ 1,  0,  0}{0.66 } & 0.34  & \textcolor[rgb]{ 1,  0,  0}{0.66 } & 0.44  & \textcolor[rgb]{ 1,  0,  0}{0.57 } & 0.46  & \textcolor[rgb]{ 1,  0,  0}{0.54 } \\
    \midrule
    \multicolumn{1}{c}{\multirow{6}[4]{*}{{\thead{Story Prediction\\Task}}}} & RECO+  & Vicuna-13B & \textcolor[rgb]{ 1,  0,  0}{0.95 } & 0.05  & \textcolor[rgb]{ 1,  0,  0}{0.70 } & \textcolor[rgb]{ 0,  .439,  .753}{0.31 } & \textcolor[rgb]{ 1,  0,  0}{0.81 } & 0.19  & \textcolor[rgb]{ 1,  0,  0}{0.99 } & 0.02  \\
          & LLaVa+ & Vicuna-13B & \textcolor[rgb]{ 1,  0,  0}{0.77 } & \textcolor[rgb]{ 0,  .439,  .753}{0.23 } & \textcolor[rgb]{ 1,  0,  0}{0.73 } & 0.27  & \textcolor[rgb]{ 1,  0,  0}{0.76 } & 0.24  & \textcolor[rgb]{ 1,  0,  0}{0.72 } & 0.28  \\
          & MiniGPT-5 & Vicuna-7B  & \textcolor[rgb]{ 1,  0,  0}{0.97 } & 0.03  & \textcolor[rgb]{ 1,  0,  0}{0.70 } & 0.30  & \textcolor[rgb]{ 1,  0,  0}{0.69 } & \textcolor[rgb]{ 0,  .439,  .753}{0.31 }  & \textcolor[rgb]{ 1,  0,  0}{0.53 } & \textcolor[rgb]{ 0,  .439,  .753}{0.47 } \\
\cmidrule{2-11}          & LLaMS-7B & Vicuna-7B  & 0.54  & 0.46  & 0.54  & 0.46  & 0.51  & 0.49  & 0.51  & 0.49  \\
          & G.T.  & -     & 0.47  & \textcolor[rgb]{ 1,  0,  0}{0.53 } & 0.42  & \textcolor[rgb]{ 1,  0,  0}{0.58 } & 0.31  & \textcolor[rgb]{ 1,  0,  0}{0.69 } & 0.47  & \textcolor[rgb]{ 1,  0,  0}{0.54 } \\
    \bottomrule
    \end{tabular}}%
  \label{tab:result_2}%
\end{table*}%

\subsection{Consistent Vision Storytelling Model}
With story plots predicted in the textual storytelling stage, a vivid story also requires consistent visual illustrations correlated to text contents, as shown in Fig.~\ref{fig:data} (right). In this work, we present a parameter-efficient fine-tuning module, called SQ-Adapter, to capture implicit variable-length sequence image styles and guide the story illustration generation.

The details of SQ-Adapter are illustrated in Fig.~\ref{fig:sq-adapter}. We set a learned latent query $s \in \mathbb{R}^{L_s \times C}$ as an image sequence style learner. For observed image sequence $i_{0:k}$, we capture observed visual features $\tau_{0:k}$ by $\tau_{0:k}=CLIP_I \left(i_{0:k} \right)$, then concatenate the fourier position embedding $p_{0:k}$ in feature dimension and obtain new features by $\overline{\tau}_{0:k}=\left[\tau_{0:k};p_{0:k}\right]$. To unify the common information in multiple images, we build the model with two components: (i) a cross-attention module that captures visual information by the correlation between query $s$ and visual features, and (ii) a multiheaded self-attention ($MSA$) that captures internal associations of information. The cross-attention module and $MSA$ are built in alternation, and followed by a feed-forward network (FFN):
% \begin{equation}
% \operatorname{Attention}(Q, K, V)=\operatorname{softmax}\left(\frac{Q K^T}{\sqrt{d}}\right) \cdot V,
% \end{equation}

% \begin{equation}
% \begin{split}
% s_{j-1}=s_{j-1}+\operatorname{Atten}&(s_{j-1}, \tau_{0:k})=s_{j-1}+\operatorname{softmax}\left(\frac{Q K^T}{\sqrt{d}}\right) \cdot V\\
% Q=W_Q \cdot s_{j-1}&, K=W_K \cdot \tau_{0:k}, V=W_V \cdot \tau_{0:k} \\
% s_j &= s_{j-1}+MSA\left(s_{j-1}\right).
% \end{split}
% \end{equation}

\begin{equation}
\begin{split}
s_{j-1}&=\operatorname{FFN}\left(s_{j-1}+\operatorname{Atten}(s_{j-1}, \overline{\tau}_{0:k})\right)+s_{j-1} \\
s_j &= \operatorname{FFN}\left(s_{j-1}+MSA\left(s_{j-1}\right)\right)+s_{j-1}, \\
&\operatorname{Atten}(s_{j-1}, \overline{\tau}_{0:k}) = \operatorname{softmax}\left(\frac{Q K^T}  
{\sqrt{d}}\right) \cdot V\\
Q&=W_Q \cdot s_{j-1}, K=W_K \cdot \overline{\tau}_{0:k}, V=W_V \cdot \overline{\tau}_{0:k} .
\end{split}
\end{equation}
$s_0$ is sampled from $\mathcal{N}\left(0,1\right)$.

With the image style information, we change the SD model by an effective and lightweight adapter structure to control image generation apart from text. The rest architecture remains fixed. During the denoising process of SD, U-Net $\epsilon_\theta$ refers to the learned latent query $s$ by a trainable cross-attention module at each step. The referring information from text and vision are added and provided to $\epsilon_\theta$. The new diffusion process can be represented by
\begin{equation}
x_{t-1}=\frac{1}{\sqrt{\alpha_t}}\left(x_t-\frac{1-\alpha_t}{\sqrt{1-\bar{\alpha}_t}} \epsilon_\theta\left(x_t, c, s, t\right)\right)+\sigma_t z.
\end{equation}
Based on our architecture, SQ-Adapter can flexibly integrate image features and reason image styles for consistent story illustration generation without making assumptions specific to image length.
% , $\boldsymbol{\epsilon} \sim \mathcal{N}(0, \mathbf{I})$ is the sampled noise from Gaussian noise, $t \in \left[0,T\right]$ means the denoise step of diffusion process. 
% use a learnable latent query $q$ as an initial feature. The latent query uses a Cross Attention model $A_{multimodal}$ to capture variable-length time series information $\Bar{q}$, including observed story image features and corresponding text features, $\Bar{q} = A_{multimodal}(q, \boldsymbol{v}_{<k}, \boldsymbol{y}_{<k})$. Both image features $\boldsymbol{v}_{<k}$ and text features $\boldsymbol{y}_{<k}$ are extracted by the CLIP model. The length of the learnable query is a hyperparameter which is set much smaller than the sequence multimodal feature.
% After information fusion, the latent query will be sent to the latent transformer for further information fusion, $\widehat{q} = A_{self}(\Bar{q})$. The final query $\widehat{q}$ contains the style and story information of the historical picture. Similar to the text prompt controlling way, we use another cross attention $A_{style}$ to fuse the style information into the denoised U-Net in a stable diffusion model. 
% In the consistent vision generation stage, we train the SQ-Adapter model $\boldsymbol{\epsilon}$ by 
% SQ-Adapter is trained by generating the $k$-th image based on the $k$-th text story and images before $k$-th in the image sequence, with the training object set as follows:

\section{Experiment}
% In this section, we will present human evaluation results, visual examples, and useful applications of our approach.

\subsection{Experimental Setup}
\subsubsection{Dataset}
VIST is the most used benchmark dataset in the storytelling task. VIST comprises 210,819 unique photos sourced from 10,117 Flickr albums. The dataset is manually annotated with a series of storylines based on scene clips. Following previous works~\cite{Yang2019KnowledgeableSA,hu20aaai}, we set the storytelling task to a fixed story length of 5. 
% During story generation, methods generate story text with 5 images only. In the story prediction, methods predict future 2 images and 5 text story segments with 3 images only.

\subsubsection{Baseline Models.}
% In previous storytelling task, AREL~\cite{wang2018no} and RECO~\cite{hu20aaai} dedicate to the story generation. MiniGPT-5~\cite{zheng2023minigpt5} deals with story prediction. Therefore, we evaluate story generation and story prediction tasks, separately. 
We first evaluate the story generation task from the whole image sequence to the story text, compared with previous storytelling models (AREL~\cite{wang2018no}, RECO~\cite{hu20aaai}) and recent LLM-based image-to-text models (LLaVa~\cite{liu2023llava}, NextGPT\cite{wu2023nextgpt}). Secondly, we evaluate the multi-modal story prediction task from part of image sequence to the whole story with images and texts, compare with the combined pipelines (RECO+Vicuna-13B+SD, LLaVa+Vicuna-13B+SD) and multi-modal prediction LLM (MiniGPT-5~\cite{zheng2023minigpt5}).
% Our approach is a unified framework for both story generation and story prediction. However, existing methods focus on different tasks, they are hard to accomplish in a unit architecture. Therefore, we evaluate story generation and story prediction tasks, separately. 
% We adopt both automatic metrics and human evaluation metrics to evaluate the impressiveness and consistency of different methods. Four automatic metrics are used in our experiments: CLIP-based metrics~\cite{rombach2022high},  Inception Score (IS), and Frechet Inception Distance (FID). 

\subsubsection{Evaluation Metrics}
Automatic metrics often struggle to evaluate the expressiveness and consistency of long text stories. In this work, we leverage human evaluation to validate the effectiveness of our pipeline. 
We use comparisons between different models and our proposed LLaMS-13B to gauge the performance of all models. 100 randomly selected stories from the VIST test dataset are sent to crowdsourcing and 
each comparative evaluation is assigned to 3 workers to relieve reading variance. To eliminate inertial selection cross samples, we shuffle the order of the options in each sample. 
% they are asked to choose the method that performed better or choose a tie.

We perform human evaluation studies for the story generation task in four metrics. \textbf{Integrality} measures the basic elements of the story: beginning, end, and the protagonist. \textbf{Interestingness} measures the sustained interest in reading driven by story content. \textbf{Consistency} measures reasonableness and continuity of the storyline across multiple text plots and image sequence. In story generation task, we only evaluate consistency between text plots.
\textbf{Correlation} measures the content and order of the text being the same as the content and order of the image. 
% Besides, in the story prediction task, we also evaluate \textbf{Consistency-Image} which reasonableness and continuity of the storyline across predicted images and observed images.

\subsubsection{Implementation Details}
We utilize the LLaVa~\cite{liu2023llava} as our LLaMS base model for multi-modal alignment. The image encoder uses the ViT~\cite{dosovitskiy2020image} and the LLM uses the Vicuna~\cite{vicuna2023}. For the LLM-preferred image token, we set 32 tokens with a dimension of 768 for each image. $\lambda_g$ and $\lambda_p$ are both set to 1. In the SQ-Adapter model, the denoising U-Net is trained by SD v1.5. The latent query length is set to 4 with a dimension of 768, the same as the hidden dimension of CLIP text embedding. The depth of cross attention and latent transformer are set to 4. The same to SD, LLaMS drop text, image, and text-image pair individually with a probability of 0.05.

\begin{figure*}[ht]
\begin{center}
\includegraphics[width=17.5cm]{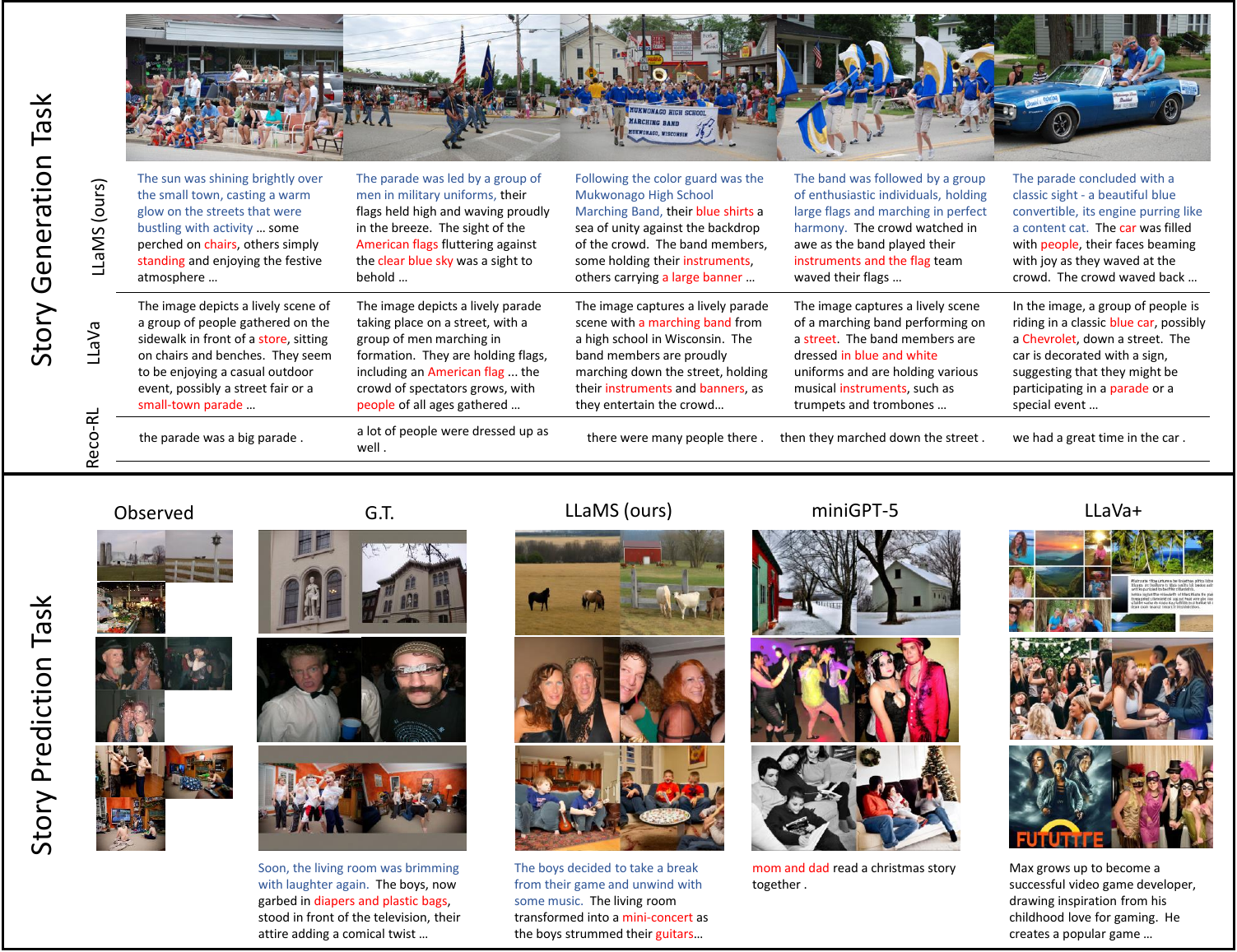}
\end{center}
\caption{The qualitative results of story generation task (Given 5 images) and story prediction task (Given 3 images). To save space, we omit the textual results for some samples in the story prediction task, and we will show more complete results in the supplementary.}
\label{fig:5to5result}
\end{figure*}

\subsection{Results on VIST}
\subsubsection{Story generation task}
We can see from Tab.~\ref{tab:result_2} that the non-LLM models(AREL and RECO) perform extremely poorly in terms of Correlation (more than 90\% Lose rate) and interest (100\% Lose rate). The numerical results verify that the non-LLM methods seriously lack expressive ability. As a general multimodal LLM, LLaVa significantly improves the Correlation compared to non-LLM methods. However, the performance over other metrics such as Interestingness and Consistency (100\% Lose rate) are still unsatisfactory. Although the general LLMs achieve basic modal alignments, they cannot make readable storytelling. Nextgpt is capable of understanding multiple images, it improves Interestingness and Consistency of the telling story (around 10 \% win rate). In order to make a fair comparison with Nextgpt, we present the performance of LLaMS-7B (using Vicuna-7B as LLM) in Tab.~\ref{tab:result_2}. The performance of our LLaMS-13B slightly surpassed that of LLaMS-7B, and significantly surpassed the performance of Nextgpt. Therefore, even with the same LLM, our LLaMS-7B still greatly improves the performance of story generation, verifying the effectiveness of our proposed pipeline. In addition, we also compared the effects of LLaMS-13B and ground truth. The experimental results showed that although LLaMS greatly improved the effect of story generation, there is still a certain gap between it and the enhanced story with the G.T. storyline.

%  \ding{52}
\begin{table*}[htbp]
  \centering
  \caption{Ablation study of sequence data enhancement. We evaluate the effect of caption data generated by a pre-trained model and storylines annotated by humans. ``w/o data enhancement" means original ground truth data in VIST dataset.}
    \scalebox{0.9}{
    \begin{tabular}{cccccccccccc}
    \toprule
    \multicolumn{1}{c}{\multirow{2}[4]{*}{\thead{method\\(data enhancement vs. *)}}} & \multirow{2}[4]{*}{caption data} & \multirow{2}[4]{*}{story lines} & \multirow{2}[4]{*}{GPT} & \multicolumn{2}{c}{Integrality} & \multicolumn{2}{c}{Interestingness} & \multicolumn{2}{c}{Correlation} & \multicolumn{2}{c}{Consistency} \\
\cmidrule{5-12}          &       &       &       & Win+  & Lose+ & Win+  & Lose+ & Win+  & Lose+ & Win+  & Lose+ \\
    \midrule
    w/o data enhancement &       & \ding{52}     &       & \textbf{0.92} & 0.09  & \textbf{0.99} & 0.01  & \textbf{0.57} & 0.43  & \textbf{0.92} & 0.08 \\
    storyline enhancement &       & \ding{52}     & \ding{52}     & \textbf{0.91} & 0.09  & \textbf{0.90} & 0.10  & \textbf{0.68} & 0.32  & \textbf{0.89} & 0.11 \\
    caption enhancement & \ding{52}     &       & \ding{52}     & \textbf{0.63} & 0.37  & \textbf{0.60} & 0.40  & \textbf{0.63} & 0.38  & \textbf{0.59} & 0.41 \\
    \bottomrule
    \end{tabular}}%
  \label{tab:data_abla}%
\end{table*}%

\begin{figure*}[ht]
\begin{center}
\includegraphics[width=\textwidth]{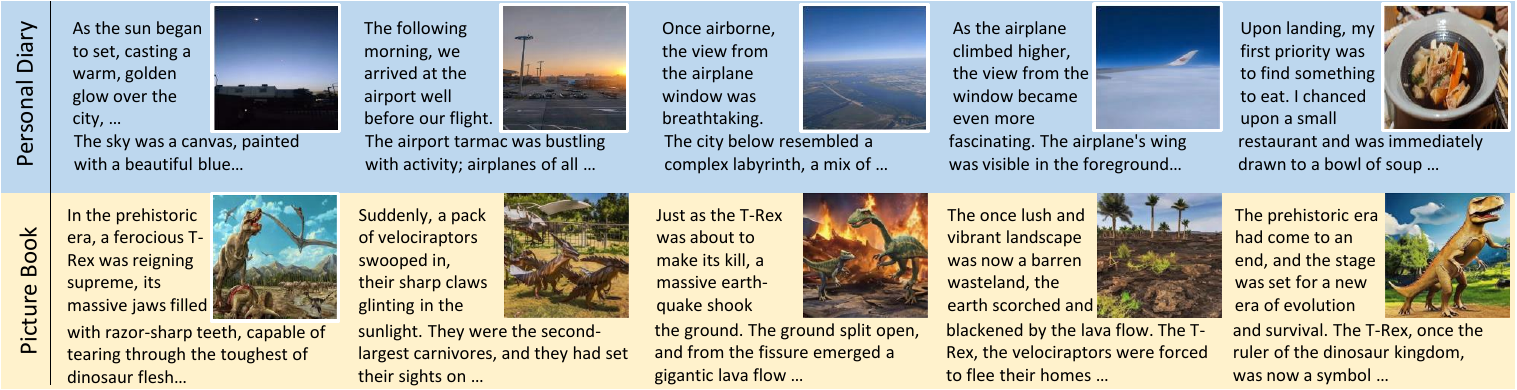}
\end{center}
\caption{Three example Applications of our LLaMS. These examples demonstrate the flexible application of our framework by integrating story generation and story prediction.}
\label{fig:result_applications}
\end{figure*}

% \begin{figure}[ht]
% \begin{center}
% \includegraphics[width=8.5cm]{figures/fig_8.pdf}
% \end{center}
% \caption{Ablation study about SQ-Adapter.}
% \label{fig:ablation_sq}
% \end{figure}
% % 这里考虑加个zero shot的case图

% Table generated by Excel2LaTeX from sheet 'ablation'
\begin{table}[htbp]
  \centering
  \caption{Ablation study for each component of LLaMS.}
    \scalebox{0.82}{\begin{tabular}{ccccccc}
    \toprule
    \multicolumn{1}{c}{\multirow{2}[4]{*}{\thead{Model\\(LLaMS-13B vs.*)}}} & \multicolumn{2}{c}{Interestingness} & \multicolumn{2}{c}{Consistency} & \multicolumn{2}{c}{Correlation} \\
\cmidrule{2-7}          & Win+  & Lose+ & Win+  & Lose+ & Win+  & Lose+ \\
    \midrule
    w/o data enhanc & \textbf{1.00 } & 0.00  & \textbf{0.86 } & 0.14  & \textbf{0.56 } & 0.44  \\
    w/o sq-adapter & -     & -     & \textbf{0.60 } & 0.40  & \textbf{0.55}     & 0.45 \\
    LLM 7B & 0.54  & 0.46  & 0.51  & 0.49  & 0.51  & 0.49  \\
    \bottomrule
    \end{tabular}}%
  \label{tab:ablation_ele}%
\end{table}%

\subsubsection{Story Prediction task}
We compare LLaMS with three-stage combined pipelines and an end-to-end story prediction model. LLaVa+ and RECO+ mean to use their original storytelling model, then rely on Vicuna-13B to predict the future two plots, and finally leverage stable diffusion to get the predicted story image. After the past processing of Vicuna, compared with their original model in story generation task, these pipelines perform well in story Consistency, but are poor in Correlation. MiniGPT-5 is a trainable large-scale multimodal architecture with 7B parameters. It promises high consistency (31$\%$ Win) but is restricted in performance in Integrality (3$\%$ Win) and Interestingness (30$\%$ Win). Our model performs best on all four metrics evaluated by humans, demonstrating our work tells human-preferred narratives.

\subsubsection{Visualization}
We show a sample story in Fig.~\ref{fig:5to5result} (top). In the story ``Celebration Parade", our LLaMS elegantly integrates fragments of different scenes to tell a vivid story. LLaVa generates the correct image content but struggles to connect it to a complete story. RECO just provides short, brief descriptions for each image. In Fig.~\ref{fig:5to5result} (bottom), we show three examples of the story prediction task. In the third example, our LLaMS can continue the storyline that happened in the observed three images and generate a story with a consistent style. MiniGPT5 failed to continue the display style, and its textual stories have less integrality and interestingness. LLaVa+ utilizes basic SD as the generator, denoising without historical images as a reference, making each image have a unique style. These examples showcase the impressive performance of our method in expressiveness and consistency.

\subsection{Ablation Study}

\textbf{Sequence data enhancement.} As shown in Tab.~\ref{tab:data_abla}, we compare the story data between different settings which is decisive guidance for storytelling methods. Compared with the original VIST data, using GPT to enrich the storyline can improve the Consistency and Interestingness slightly. While it destroys the Correlation between images and texts (drop around 10 $\%$). Using GPT to refine the caption data generated by LLaVa shows improved expressiveness and consistency. However, it is worse than our sequence data enhancement (around 20$\%$ Lose in each metric). These results demonstrate the reasonability of our sequence data enhancement strategy.

% Compared with the original data, which uses human-tagged storyline in the VIST dataset, our data enhancement is superior across all story metrics. When using GPT to refine the story data in VIST, the consistency of text data improves. If we use detailed descriptions as the source of story information, GPT can mine human-preferred stories from the descriptions. But it is still weaker than the final sequence enhanced data using VIST storyline.
% Besides, without storylines as input, the enhanced data performs badly in integrality. Both storyline enhancement and caption enhancement have high consistency in text because of the integration of GPT.

% \textbf{SQ-Adapter.} As shown in Tab.~\ref{tab:ablation_sq}, we compare the sq-adapter structure, IP-adapter structure, and finetuned stable diffusion results. We can see that the sq-adapter captures the whole style in observed three images that can generate a consistent story illustration.

\textbf{LLaMS Components.} We additionally conduct ablations of different components of LLaMS in Tab.~\ref{tab:ablation_ele}. Without enhanced data, the ablated model performs poorly in Interestingness (100 $\%$ Lose) and Consistency (86 $\%$ Lose), which demonstrates that data enhancement effectively improves the quality both in expression and consistency. When we use SD to replace the SQ-Adapter, Lose samples rise both in Consistency and Correlation. This supports our SQ-Adapter successfully capturing image sequence style and guiding human-prefer image generation. Besides, the results between 13B LLM and 7B LLM perform slightly fluctuation, indicateing that the scale of LLM has less impact on storytelling.

% Consistency-T drops to 0 compared with LLaMS results. 
% our architecture has difficulty generating consistent results over time.
% When we use SD as our image generator, 
% the image consistency drops $11\%$ in sample comparison. This demonstrates the effectiveness of the different elements in our pipeline. Besides, the parameter scale of the LLM has little effect on the final result. 
% We evaluate the effect of sequence data enhancement, LLM parameters scales, and SQ-Adapter for multi-modal story generation. As shown in Tab.~\ref{tab:ablation_ele}, 

\subsection{Applications}
Our storytelling pipeline integrates story generation and story prediction. As we asked in the Introduction, the model can be widely generalized in zero-shot. We show two application scenarios of LLaMS using images not in the VIST dataset. 

\textbf{Personal diary.} Every day, we may picture various activities on our phones, we select 5 images in one day and require the LLaMS to write a text journal with corresponding images. One sample is shown in Fig.~\ref{fig:result_applications} (top).

% \textbf{Ending creation} When watching movies and TV shows, we often get stuck in the script written by the screenwriter. A bad ending can haunt us for days. We can use LLaMS to predict various endings. One sample is shown in Fig.~\ref{fig:result_applications}(middle).

\textbf{Picture book.} When we want to prepare a storybook for children, we presuppose the beginning of a story (1 image), and may be hard to find consistent images and text to continue. We can leverage LLaMS for the rest. One sample is shown in Fig.~\ref{fig:result_applications} (bottom).

\section{Conclusion}
In our work, we propose a novel pipeline, called LLaMS, used for a vivid and reasonable story generation. We present an automatic sequence data enhancement to create high-quality story data, based on pre-trained multi-modal LLM and language-only LLM, improving the expressiveness of our architecture. We also present the sq-adapter to maintain visual consistency across multiple images during story prediction. Our human evaluation and work applications show excellent results. The storytelling task is a sequence reasoning task, and several directions can be continued explored in future work, such as Id-Consistency for the single protagonist and Long Storytelling with tortuous plots.
% (1) Id-consistency. Current sequence data enhancement hardly records the same individual in different plots. Therefore, the storyline is randomly chosen as multi-people events or single-people events. It is worth exploring for further id-consistency.
% (2) Abundant story plots. The storyline in the VIST dataset is still limited to 10k flicker albums which is far to enough for daily storytelling. Collecting from movie-related datasets is a convenient way to explore the current dataset.
% (3) Long storytelling. Generating 5 stories limits the reasoning of the story. More plots help to create an overall story with twists and turns.

{
    \small
    \bibliographystyle{named}
    \bibliography{main}
}

% \input{X_suppl}

\end{document}

\maketitle

\begin{abstract}
In section A of this supplementary, we provide instruction designing in our work. In section B, we show the original human evaluation results and more qualitative results. We also analyze the Interclass Correlation Coefficient for human evaluation results.
In section C, we discuss the limitations of our work and future directions for exploration.
\end{abstract}

\section{Instruction Designing}
As we leverage pre-trained LLM for sequence data augmentation and visual content understanding, instruction designing greatly affects the generated high-quality data and storytelling results. We present our Instructions used for different stages.

\begin{figure*}[ht]
\begin{center}
\includegraphics[width=15cm]{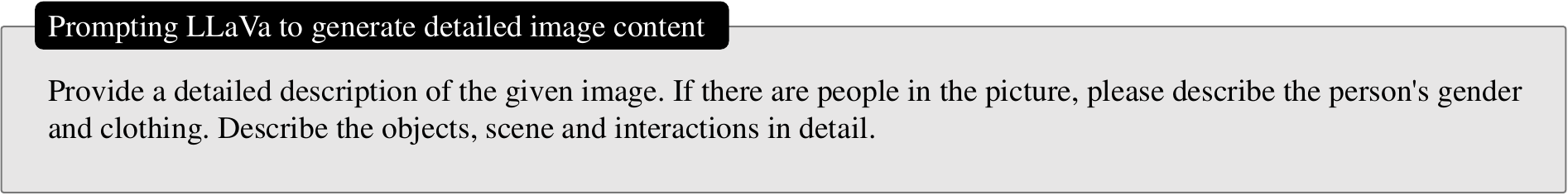}
\end{center}
\caption{The instruction for image detailed description using pre-trained LLaVa.}
\label{tab:llava}
\end{figure*}

\textbf{Instruction for detailed image description.} During the sequence data enhancement stage, we design an elaborate instruction used to describe the image content in detail using LLaVa~\cite{liu2023llava}, as shown in Table~\ref{tab:llava}. A description that is highly relevant to the images and detailed in content can provide useful information for subsequent story rewrite.

\begin{figure*}[ht]
\begin{center}
\includegraphics[width=15cm]{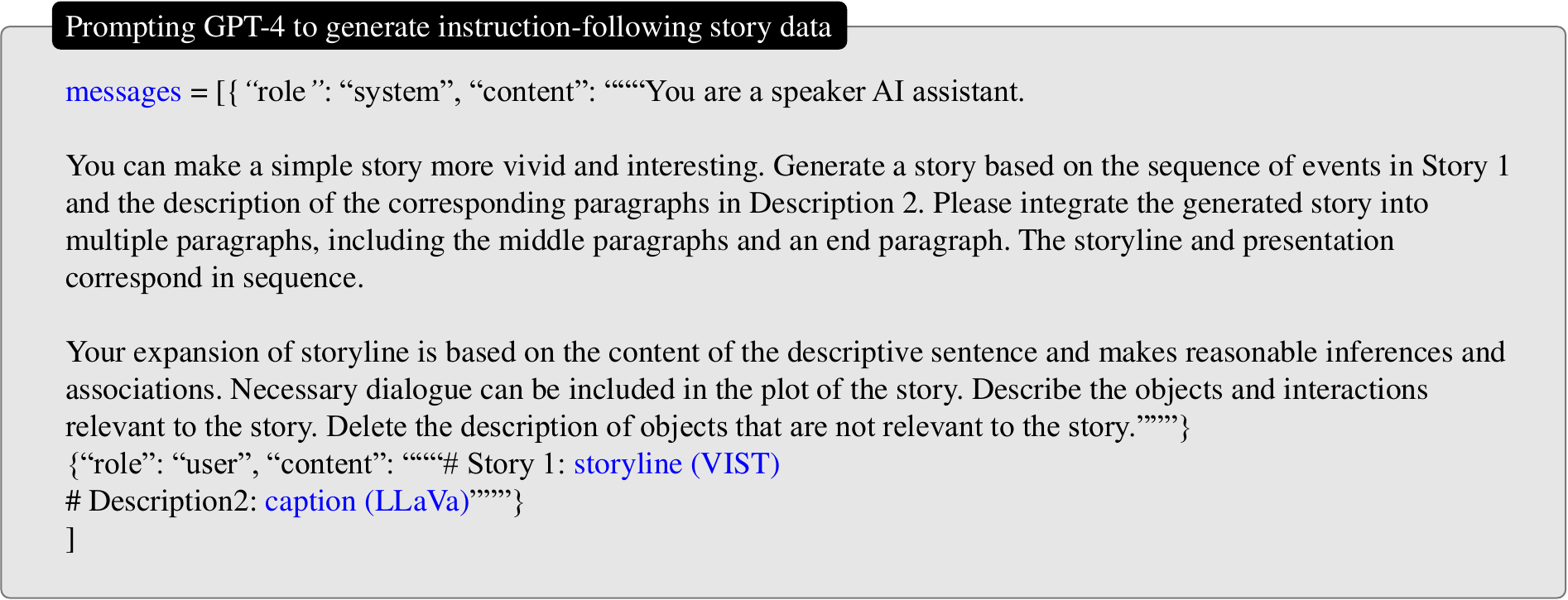}
\end{center}
\caption{We prompt the GPT-4 to generate an integrated story, keeping the original storyline and imagining the plot content based on corresponding descriptions.}
\label{tab:gpt}
\end{figure*}

\textbf{Instruction for vivid stories enhancement.}  With the original storyline in VIST and detailed descriptions generated by LLaVa, we prompt the GPT-4 to generate an integrated story, keeping the original storyline and imagining the plot content based on corresponding descriptions. The instructions are shown in Table~\ref{tab:gpt}.

\begin{figure*}[ht]
\begin{center}
\includegraphics[width=15cm]{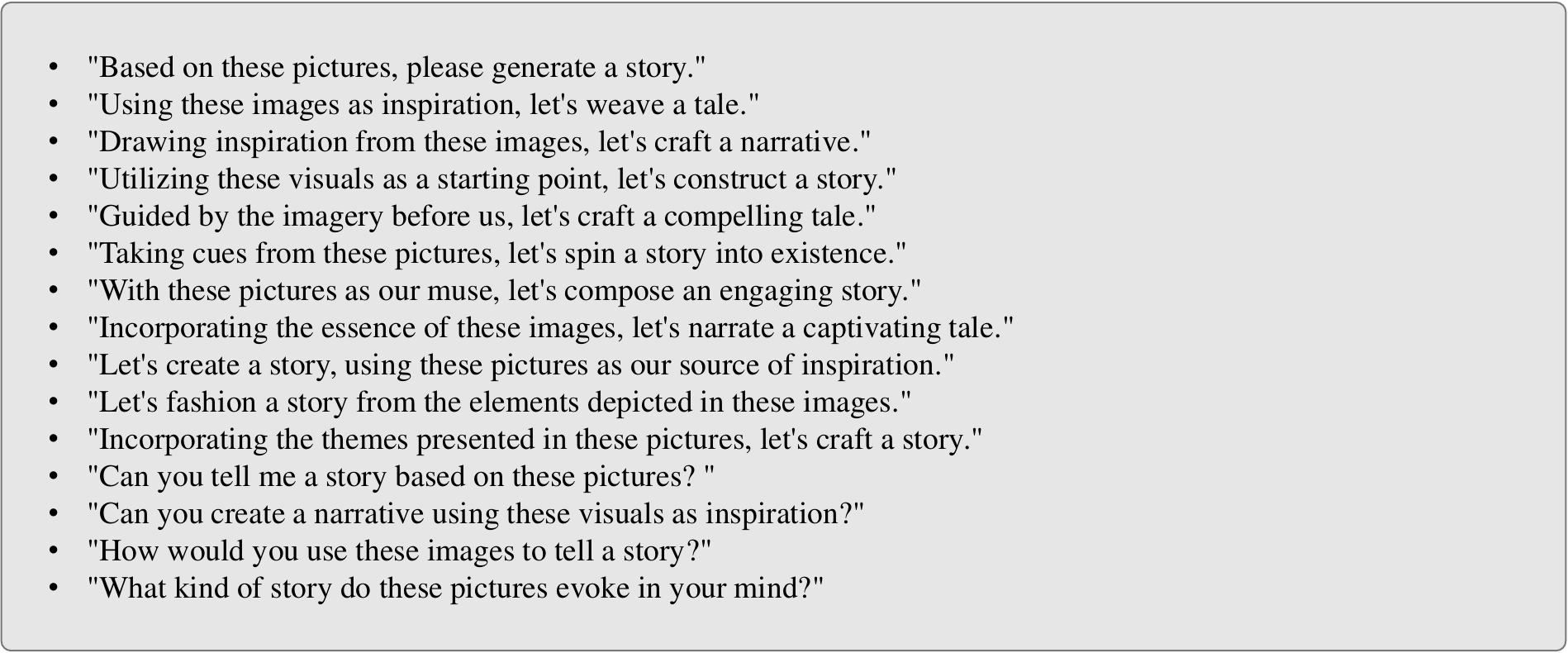}
\end{center}
\caption{The list of instructions for storytelling.}
\label{tab:instructions}
\end{figure*}

\textbf{Instructions for storytelling.} In our LLaMS, we provide the instruction list for storytelling, including story generation task and story prediction task, shown in Table~\ref{tab:instructions}. These instructions present the same meaning with variant sentences.

\section{Experimental Results}
We evaluate the story generation task from the whole image sequence to the story text, compared with previous storytelling models (AREL~\cite{wang2018no}, RECO~\cite{hu20aaai}) and recent LLM-based image-to-text models (LLaVa~\cite{liu2023llava}, NextGPT\cite{wu2023nextgpt}). Secondly, we evaluate the multi-modal story prediction task from part of image sequence to the whole story with images and texts, compare with the combined pipelines (RECO+Vicuna-13B~\cite{vicuna2023}+SD~\cite{rombach2022high}, LLaVa+Vicuna-13B+SD) and multi-modal prediction LLM (MiniGPT-5~\cite{zheng2023minigpt5}).
\subsection{Qualitative Results}
\begin{figure*}[ht]
\begin{center}
\includegraphics[width=17.5cm]{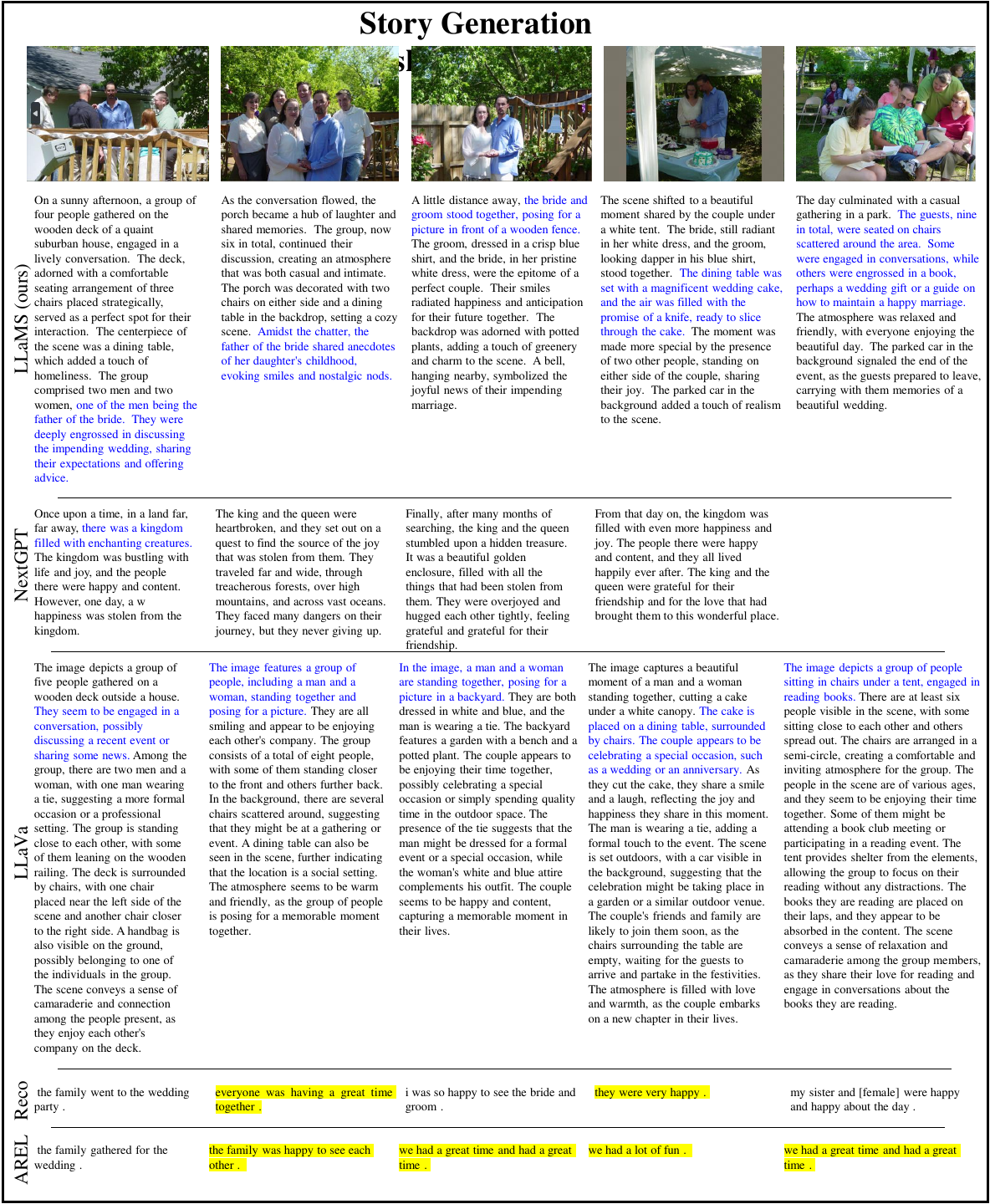}
\end{center}
\caption{Complete sample 1 of different methods in story generation task (Given 5 images).}
\label{fig:generation_1}
\end{figure*}

\begin{figure*}[ht]
\begin{center}
\includegraphics[width=17.5cm]{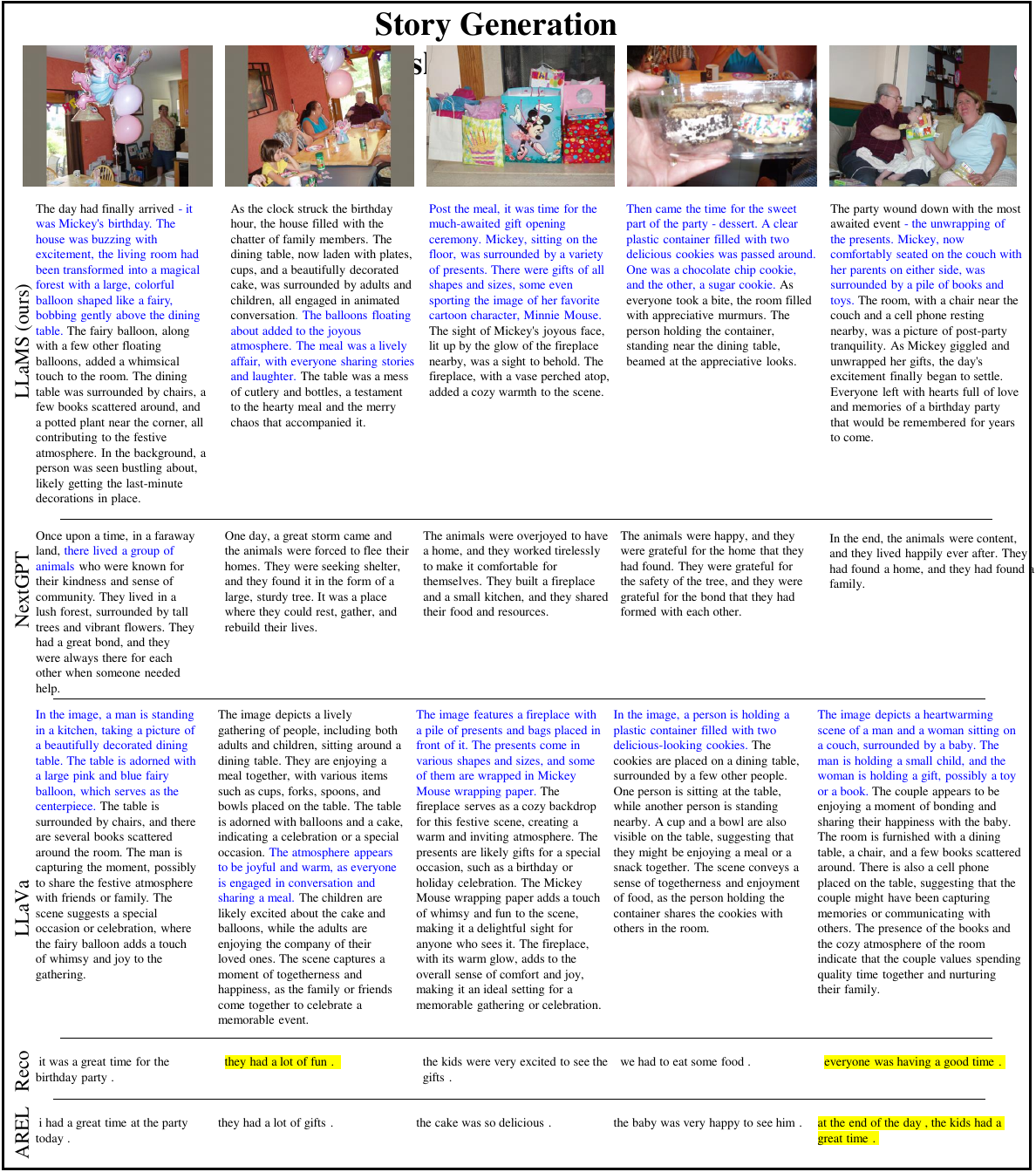}
\end{center}
\caption{Complete sample 2 of different methods in story generation task (Given 5 images).}
\label{fig:generation_2}
\end{figure*}

\begin{figure*}[ht]
\begin{center}
\includegraphics[width=17.5cm]{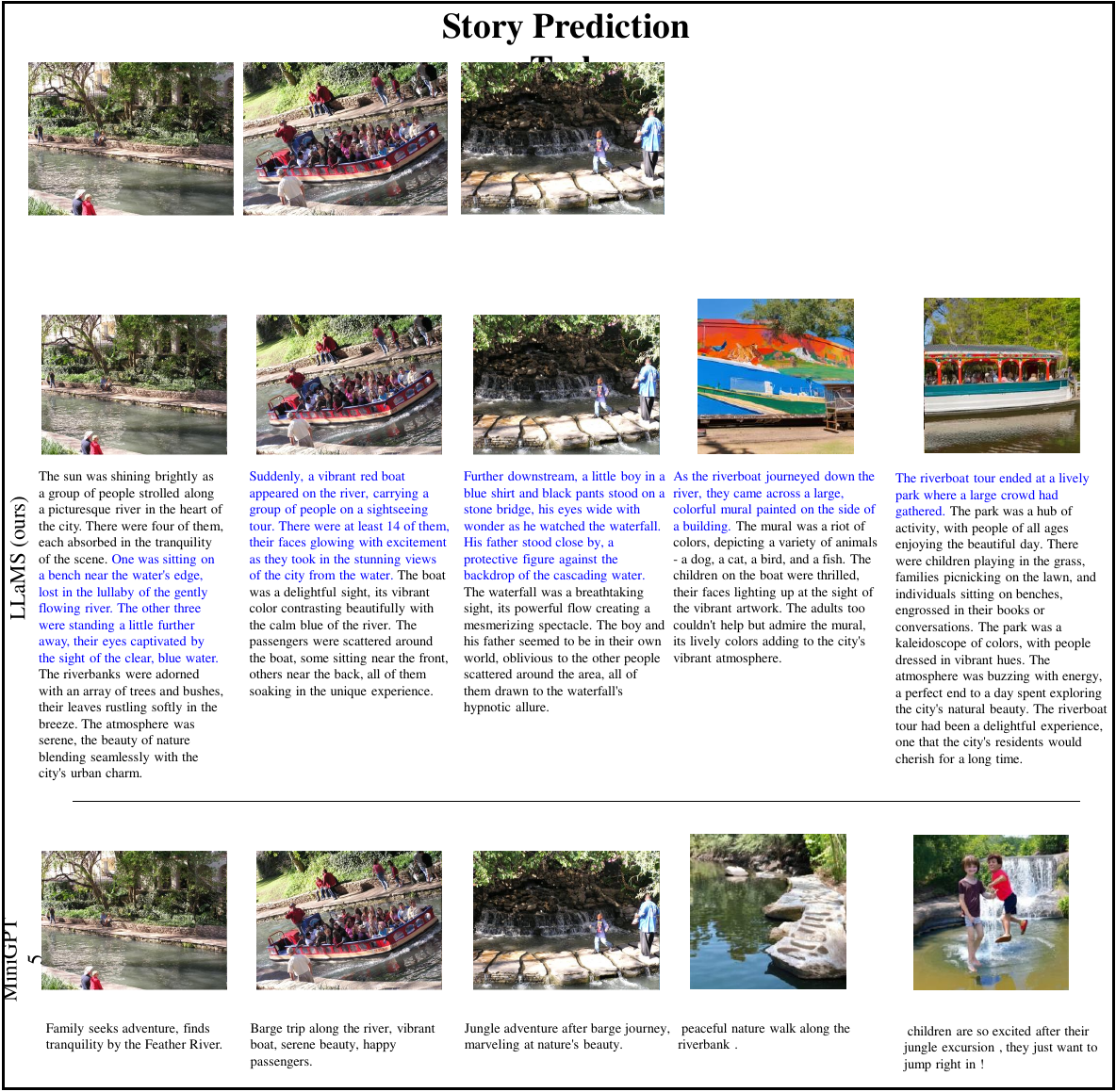}
\end{center}
\caption{Complete sample 1 of different methods (ours, MiniGPT5) in story prediction task (Given 3 images).}
\label{fig:prediction_10}
\end{figure*}

\begin{figure*}[ht]
\begin{center}
\includegraphics[width=17.5cm]{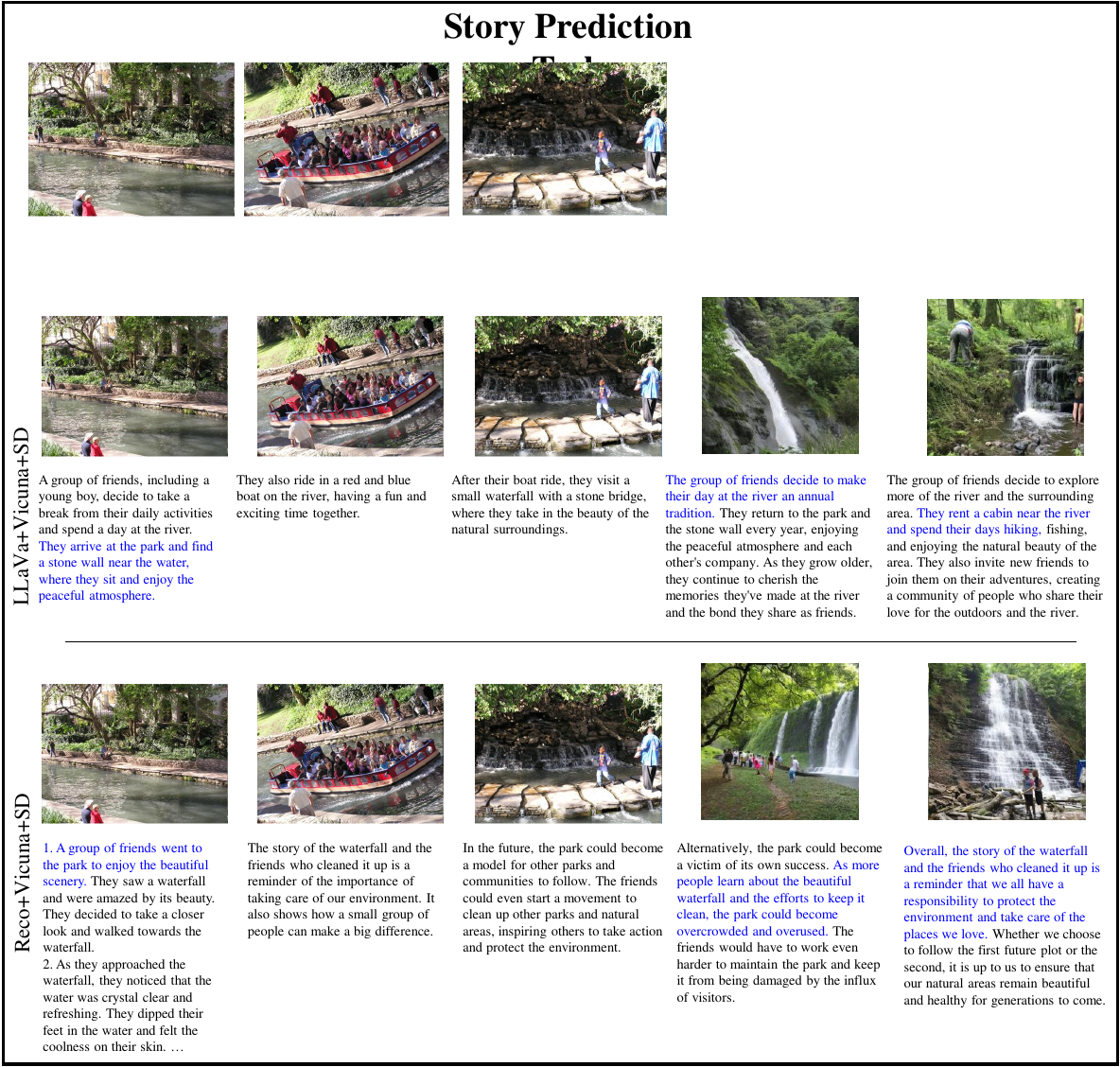}
\end{center}
\caption{Complete sample 1 of different methods (LLaVa+Vicuna+Stable Diffusion, Reco+Vicuna+Stable Diffusion) in story prediction task (Given 3 images).}
\label{fig:prediction_11}
\end{figure*}

\begin{figure*}[ht]
\begin{center}
\includegraphics[width=17.5cm]{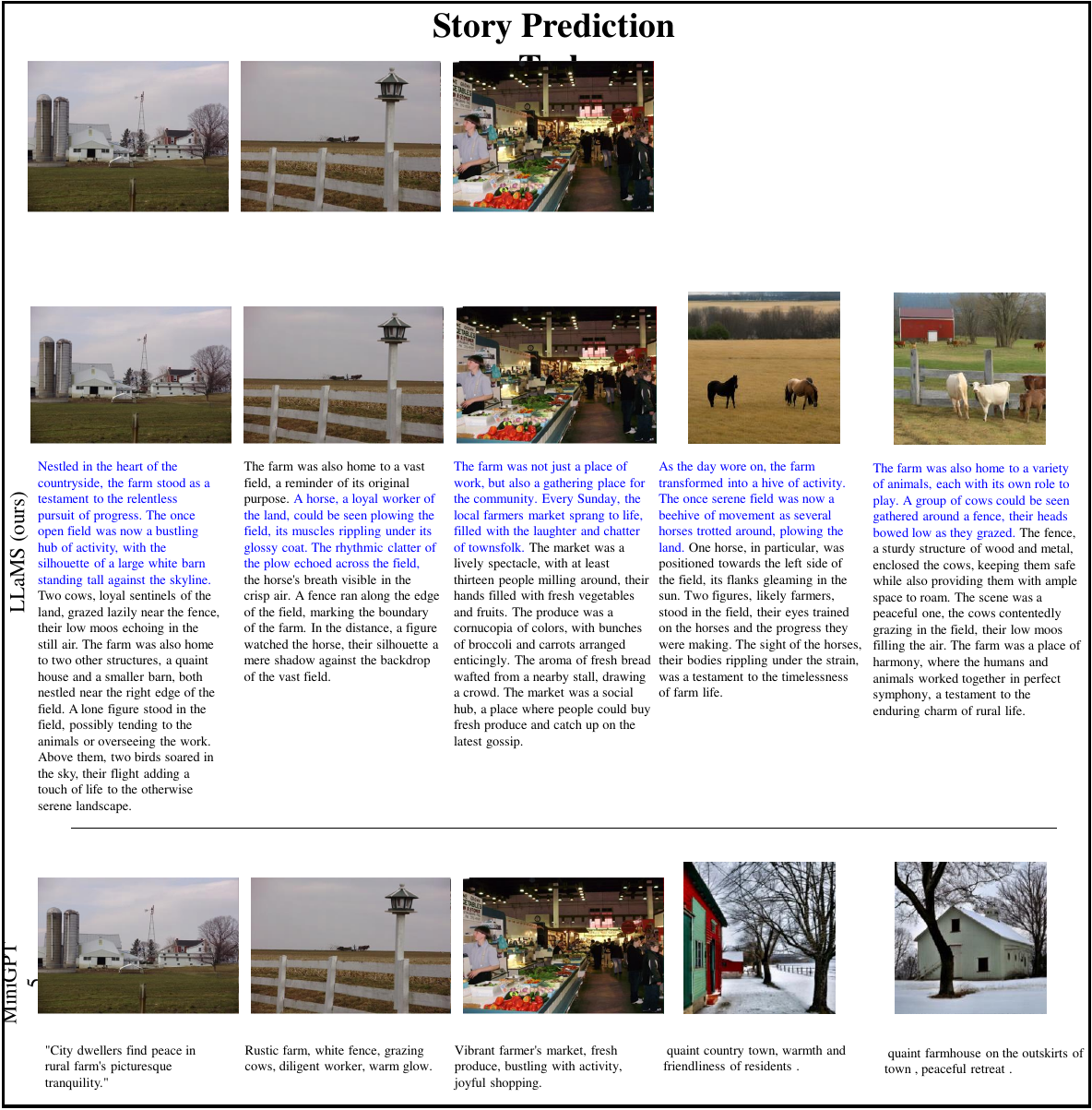}
\end{center}
\caption{The qualitative results of different methods (ours, MiniGPT5) in story prediction task (Given 3 images).}
\label{fig:prediction_20}
\end{figure*}

\begin{figure*}[ht]
\begin{center}
\includegraphics[width=17.5cm]{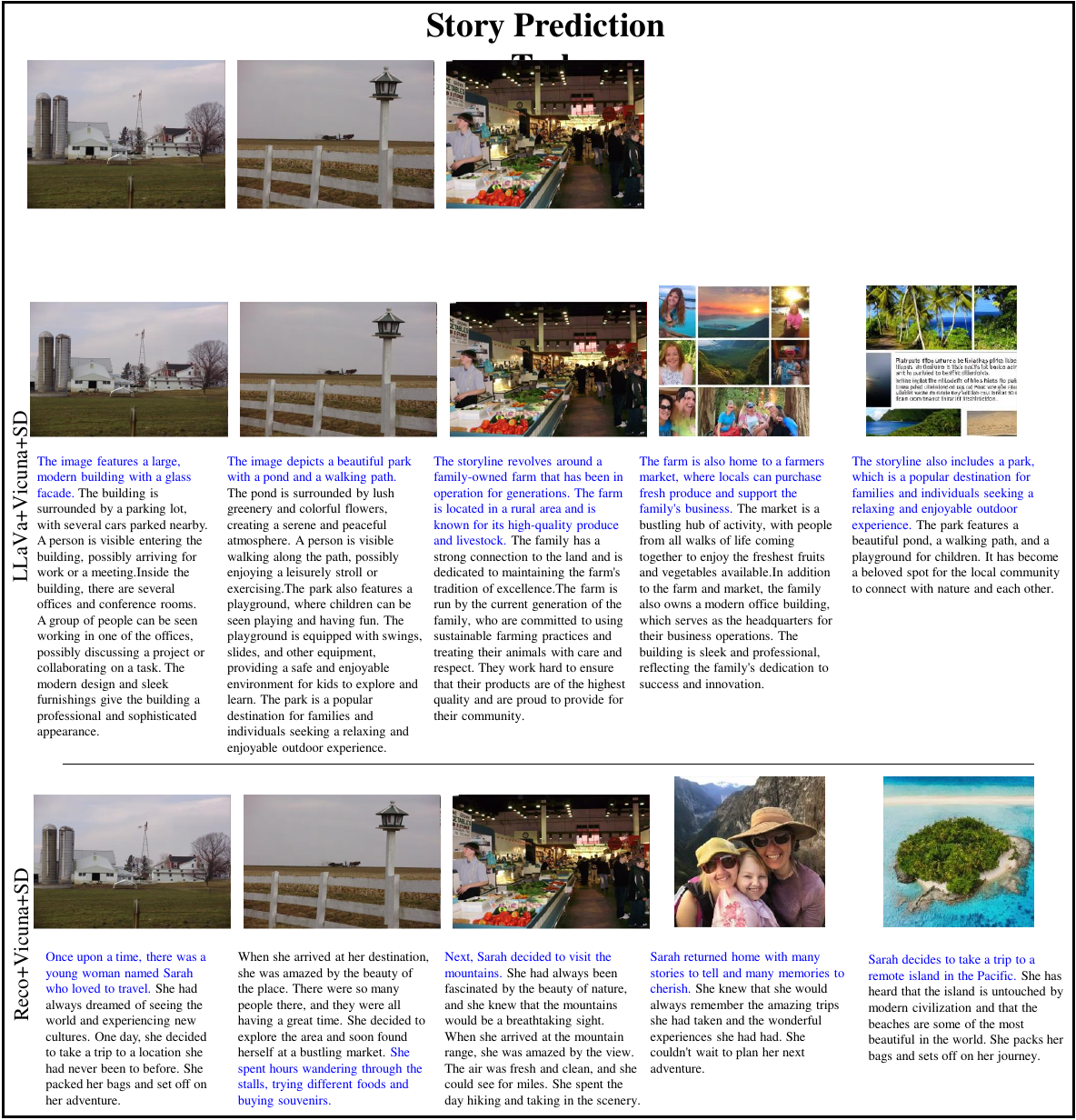}
\end{center}
\caption{The qualitative results of different methods (LLaVa+Vicuna+Stable Diffusion, Reco+Vicuna+Stable Diffusion) in story prediction task (Given 3 images).}
\label{fig:prediction_21}
\end{figure*}

We show two more complete examples here. The examples are from VIST testing data. Story generation task examples are illustrated in Fig.~\ref{fig:generation_1}, Fig.~\ref{fig:generation_2}. Story prediction task examples are illustrated in Fig.~\ref{fig:prediction_10}, Fig.~\ref{fig:prediction_11}, Fig.~\ref{fig:prediction_20}, and Fig.~\ref{fig:prediction_21}. We highlight the storyline in long story plots by blue color, and highlight meaningless story descriptions by yellow underpainting. 

Take the Fig.~\ref{fig:generation_1} as an example, our LLaMS recognize the storyline as ``wedding". Although the bride is just wearing a white shirt, not a long wedding dress, and not every picture can be seen to be related to the wedding. LLaMS tells the story by ``discussing the impending wedding", ``bride father share daughter's childhood", ``posing for a picture", ``slice through the cake", and ``engrossed in a book" in each plot. This demonstrates that our approach can jointly understand the content of multiple images and generate sequence-consistent story plots with a whole storyline throughout the story. NextGPT~\cite{wu2023nextgpt} observes five images simultaneously, but the story is less correlated to the content of the images. Because the generated story is limited by the knowledge of LLM about storylines, which prefers ``adventure" and lacks diversity. LLaVa~\cite{liu2023llava} describes every detail in the image, but hard to synthesize a coherent story. AREL~\cite{wang2018no} and RECO~\cite{hu20aaai} use brief sentences to tell the story, while the results are frustrating due to poor narrative. This example demonstrates that our proposed pipeline can tell a more vivid story.

\subsection{Human Evaluation Results}
% \subsubsection{Baseline Models.}
In the main paper, for clarity, we combine the win, lose, and tie as Win+Tie/2 and Lose+Tie/2. Here, we represent the raw evaluation results in Tab.~\ref{tab:1}, Tab.~\ref{tab:2}, and Tab.~\ref{tab:3}, respectively.

% \subsubsection{Evaluation Metrics}
As discussed in the main paper, we perform human evaluation studies for the story generation task in four metrics. \textbf{Integrality} measures the basic elements of the story: beginning, end, and the protagonist. \textbf{Interestingness} measures the sustained interest in reading driven by story content. \textbf{Consistency} measures reasonableness and continuity of the storyline across multiple text plots and image sequences. In the story generation task, we only evaluate consistency between text plots. 
\textbf{Correlation} measures the content and order of the text being the same as the content and order of the image. 
% Table generated by Excel2LaTeX from sheet 'results'
\begin{table}[htbp]
  \centering
  \caption{Interclass Correlation Coefficient analysis for human evaluation results.}
    \begin{tabular}{ccccc}
    \toprule
    \multirow{2}[4]{*}{Methods} & \multicolumn{4}{c}{Integrality} \\
\cmidrule{2-5}          & Win($\%$)   & \multicolumn{1}{l}{Lose($\%$)} & \multicolumn{1}{l}{Tie($\%$)} & \multicolumn{1}{l}{ICC} \\
    \midrule
    AREL  & 0.86  & 0.09  & 0.05  & \textcolor[rgb]{ .753,  0,  0}{0.82 } \\
    RECO  & 0.88  & 0.05  & 0.08  & \textcolor[rgb]{ .753,  0,  0}{0.96 } \\
    LLaVa  & 0.96  & 0.00  & 0.04  & \textcolor[rgb]{ .753,  0,  0}{0.98 } \\
    NextGPT & 0.48  & 0.24  & 0.28  & 0.64  \\
    \midrule
    LLaMS-7B & 0.42  & 0.35  & 0.24  & \textcolor[rgb]{ .188,  .329,  .588}{0.23 } \\
    G.T.  & 0.20  & 0.52  & 0.28  & \textcolor[rgb]{ .188,  .329,  .588}{0.42 } \\
    \midrule
    RECO+  & 0.95  & 0.05  & 0.01  & \textcolor[rgb]{ .753,  0,  0}{0.82 } \\
    LLaVa+ & 0.63  & 0.08  & 0.29  & 0.64  \\
    MiniGPT-5 & 0.95  & 0.00  & 0.05  & \textcolor[rgb]{ .753,  0,  0}{0.77 } \\
    \midrule
    LLaMS-7B & 0.30  & 0.22  & 0.48  & \textcolor[rgb]{ .188,  .329,  .588}{0.08 } \\
    G.T.  & 0.34  & 0.41  & 0.25  & \textcolor[rgb]{ .188,  .329,  .588}{0.06 } \\
    \bottomrule
    \end{tabular}%
  \label{tab:icc}%
\end{table}%

\textbf{Interclass Correlation Coefficient Analysis.} In our experimental setup, we randomly select raters in Crowdsourcing. Each method requires three raters to evaluate, and raters are not shared between different methods. To measure the generalization verification, we use Interclass Correlation Coefficient~\cite{koo2016guideline} (ICC(2,k)) to calculate the absolute agreement between evaluation results. We take the story integrity evaluation metric as an example, as shown in Table ~\ref{tab:icc}. Our method outperforms previous storytelling works, such as AREL, LLaVa, and MiniGPT5, with a high winning rate and ICC values above 0.75, indicating that raters consistently believe that our method performs better. When compared with LLaMS-7B and ground truth, the statistical probabilities of winning and losing are similar. The ICC is lower than 0.5, indicating that the raters have different opinions on the selection of some samples. The ICC result is consistent with the performance of comparison, indicating that samples with similar performance are prone to opinion bias for raters. Interclass Correlation Coefficient Analysis also demonstrates the superior performance of our method.

\section{Limitations and Future Work.}
Storytelling is a task with a high degree of freedom for generated results. It only requires to be faithful to the content of observations, while the potential storylines can be diverse. Our work performs excellently and achieves state-of-the-art results in variance metrics, including integrality, interestingness, consistency, and correlation. However, we can still find some explicit limitations which are our future directions.

\textbf{Redundant Description.} Our proposed pipeline uses the enhanced data as new training data which means its performance is greatly affected by the quality of the new data. We leverage the GPT-4 to generate a new story that has the same storyline as the original dataset and enrich the plot content using detailed descriptions. However, GPT-4 hardly balances the length of story content and object description, eventually generating redundant descriptions. A reasonable direction to explore is to adjust prompts and use the few-shot instruction tuning.

\textbf{Person Identification Inconsistent.} An excellent story requires that the person identification can maintain consistency in different plots. This is a challenging direction for multimodal storytelling that requires identifying the same person among multiple images and generating a reasonable story based on its action. The same in story illustration prediction, our pipeline has difficulty in recognizing the protagonist and generating a consistent protagonist in future story images. This deserves further exploration to get a truer story.

\textbf{Hallucinated Influence.} Our work inherits the limitations of GPT-4 about its hallucination. Initially, we found that using the simple prompt for GPT-4, such as "Generate a new story based on the storyline and detailed descriptions", makes a new dataset with many hallucinated examples. We gradually refine the prompt to increase the data quality. However, the hallucinated samples still exist. In future directions, we can use rules or let experts revise or filter new data for higher quality.

% \newpage

% Table generated by Excel2LaTeX from sheet 'results'
\begin{sidewaystable}[htbp]
\vspace{20em}
  \centering
  \caption{Raw human evaluation results on story generation and story prediction tasks. This Table serves as a data supplement to Table 1 of the main paper.}
    \scalebox{0.9}{\begin{tabular}{cccccccccccccccccc}
    \toprule
    \multirow{2}[3]{*}{modal} & \multicolumn{1}{c}{\multirow{2}[3]{*}{\thead{Models\\ (LLaMS vs. *)}}} & \multicolumn{1}{c}{\multirow{2}[3]{*}{\thead{LLM\\(V-13B)}}} & \multicolumn{3}{c}{Integrality} & \multicolumn{3}{c}{interestingness} & \multicolumn{3}{c}{Consistency-Text} & \multicolumn{3}{c}{Consistency-Image} & \multicolumn{3}{c}{Correlation} \\
\cmidrule{4-18}          &       &       & Win   & Lose  & Tie   & Win   & Lose  & Tie   & Win   & Lose  & Tie   & Win   & Lose  & Tie   & Win   & Lose  & Tie \\
    \multicolumn{1}{c}{\multirow{6}[3]{*}{\thead{Story generation\\ Models}}} & AREL  & -     & 0.86  & 0.09  & 0.05  & 0.89  & 0.07  & 0.04  & 1.00  & 0.00  & 0.00  & -     & -     &       & 0.76  & 0.07  & 0.17  \\
          & RECO  & -     & 0.88  & 0.05  & 0.08  & 1.00  & 0.00  & 0.00  & 1.00  & 0.00  & 0.00  & -     & -     &       & 0.85  & 0.00  & 0.15  \\
          & LLaVa  & V-13B & 0.96  & 0.00  & 0.04  & 1.00  & 0.00  & 0.00  & 1.00  & 0.00  & 0.00  & -     & -     &       & 0.35  & 0.01  & 0.65  \\
          & NextGPT & V-7B  & 0.48  & 0.24  & 0.28  & 0.89  & 0.10  & 0.01  & 0.87  & 0.12  & 0.02  & -     & -     &       & 0.99  & 0.01  & 0.00  \\
\cmidrule{2-18}          & LLaMS-7B & V-7B  & 0.42  & 0.35  & 0.24  & 0.40  & 0.40  & 0.20  & 0.39  & 0.37  & 0.23  & -     & -     &       & 0.47  & 0.41  & 0.13  \\
          & G.T.  & -     & 0.20  & 0.52  & 0.28  & 0.21  & 0.54  & 0.25  & 0.20  & 0.33  & 0.47  & -     & -     &       & 0.11  & 0.19  & 0.70  \\
    \midrule
    \multicolumn{1}{c}{\multirow{5}[4]{*}{\thead{Story Prediction\\ Models}}} & RECO+  & V-13B & 0.95  & 0.05  & 0.01  & 0.62  & 0.23  & 0.14  & 0.93  & 0.06  & 0.00  & 0.49  & 0.13  & 0.38  & 0.98  & 0.01  & 0.00  \\
          & LLaVa+ & V-13B & 0.63  & 0.08  & 0.29  & 0.60  & 0.13  & 0.27  & 0.68  & 0.10  & 0.23  & 0.56  & 0.09  & 0.35  & 0.57  & 0.12  & 0.31  \\
          & MiniGPT-5 & V-7B  & 0.95  & 0.00  & 0.05  & 0.69  & 0.29  & 0.02  & 0.94  & 0.02  & 0.04  & 0.14  & 0.31  & 0.55  & 0.13  & 0.08  & 0.79  \\
\cmidrule{2-18}          & LLaMS-7B & V-7B  & 0.30  & 0.22  & 0.48  & 0.28  & 0.21  & 0.51  & 0.30  & 0.27  & 0.43  & 0.26  & 0.26  & 0.48  & 0.06  & 0.04  & 0.90  \\
          & G.T.  & -     & 0.34  & 0.41  & 0.25  & 0.36  & 0.52  & 0.11  & 0.39  & 0.42  & 0.19  & 0.07  & 0.80  & 0.13  & 0.25  & 0.32  & 0.44  \\
    \bottomrule
    \end{tabular}}%
  \label{tab:1}%
  \vspace{4em}
% \end{sidewaystable}%

% % Table generated by Excel2LaTeX from sheet 'ablation'
% \begin{sidewaystable}[htp]
  \centering
  \caption{Raw human evaluation results on ablation study of sequence data enhancement. This Table serves as a data supplement to Table 2 of the main paper.}
    \scalebox{0.9}{\begin{tabular}{cccccccccccccccc}
    \toprule
    \multicolumn{1}{c}{\multirow{2}[4]{*}{\thead{method\\(data enhancement vs. *)}}} & \multirow{2}[4]{*}{caption data} & \multirow{2}[4]{*}{story lines} & \multirow{2}[4]{*}{GPT} & \multicolumn{3}{c}{Integrality} & \multicolumn{3}{c}{interestingness} & \multicolumn{3}{c}{Correlation} & \multicolumn{3}{c}{Consistency-text} \\
\cmidrule{5-16}          &       &       &       & Win   & Lose  & Tie   & Win   & Lose  & Tie   & Win   & Lose  & Tie   & Win   & Lose  & Tie \\
    \midrule
    w/o data enhancement &       &\ding{52}     &       & 0.83  & 0.00  & 0.17  & 0.99  & 0.00  & 0.01  & 0.16  & 0.02  & 0.82  & 0.85  & 0.01  & 0.14  \\
    storyline enhancement &       &\ding{52}     &\ding{52}     & 0.86  & 0.04  & 0.10  & 0.81  & 0.01  & 0.18  & 0.59  & 0.23  & 0.18  & 0.83  & 0.05  & 0.12  \\
    caption enhancement &\ding{52}     &       &\ding{52}     & 0.50  & 0.25  & 0.25  & 0.53  & 0.32  & 0.15  & 0.51  & 0.26  & 0.22  & 0.51  & 0.33  & 0.17  \\
    \bottomrule

    \end{tabular}}%
  \label{tab:2}%
    \vspace{4em}
    \centering
  \caption{Raw human evaluation results on ablation study of each element of LLaSa. This Table serves as a data supplement to Table 3 of the main paper.}
    \scalebox{0.9}{\begin{tabular}{cccccccccc}
    \toprule
    \multicolumn{1}{c}{\multirow{2}[3]{*}{\thead{Models\\ (LLaSa vs. *)}}} & \multicolumn{3}{c}{Consistency-Text} & \multicolumn{3}{c}{Consistency-Image} & \multicolumn{3}{c}{Correlation} \\
\cmidrule{2-10}          & Win   & Lose  & Tie   & Win   & Lose  & Tie   & Win   & Lose  & Tie \\
    w/o data enhanc & 1.00  & 0.00  & 0.00  & 0.61  & 0.20  & 0.19  & 0.49  & 0.37  & 0.14  \\
    s/o sq-adapter & -     & -     & -     & 0.47  & 0.27  & 0.26  & 0.36  & 0.26  & 0.38  \\
    w/o LLM 13B & 0.30  & 0.27  & 0.43  & 0.26  & 0.26  & 0.48  & 0.06  & 0.04  & 0.90  \\
    \bottomrule
    \end{tabular}}%
  \label{tab:3}%
\end{sidewaystable}%

{
    \small
    \bibliographystyle{named}
    \bibliography{main}
}